\documentclass{article} 
\usepackage[final]{colm2026_conference}

\usepackage{microtype}
\usepackage{hyperref}
\usepackage{url}
\usepackage{booktabs}
\usepackage{graphicx}
\usepackage{tabularx}
\usepackage{adjustbox}

\usepackage{lineno}
\usepackage{float}

\definecolor{darkblue}{rgb}{0, 0, 0.5}
\hypersetup{colorlinks=true, citecolor=darkblue, linkcolor=darkblue, urlcolor=darkblue}

\title{CivBench: Progress-Based Evaluation for LLMs' Strategic Decision-Making in Civilization V}

\author{
John Chen \\
University of Arizona \\
\texttt{johnchen@arizona.edu} \\
\And
Sihan Cheng \\
Northwestern University \\
\texttt{sihancheng2026@u.northwestern.edu} \\
\And
Can Gurkan \\
Northwestern University \\
\texttt{gurkan@u.northwestern.edu}
\And
Mingyi Lin \\
University of Arizona \\
\texttt{linbj1212@outlook.com} \\
}

\begin{document}

\ifcolmsubmission
\linenumbers
\fi

\maketitle

\begin{abstract}
Evaluating strategic decision-making in LLM-based agents requires generative, competitive, and longitudinal environments, yet few benchmarks provide all three, and fewer still offer evaluation signals rich enough for long-horizon, multi-agent play. We introduce CivBench, a benchmark for LLM strategists (a model under an agentic setup) in multiplayer Civilization V. Because terminal win/loss is too sparse a signal in games spanning hundreds of turns and multiple opponents, CivBench trains estimators of a state-value function, the probability that a player eventually wins given the game state. cross 307 games with 7 LLMs and multiple CivBench agent conditions, we demonstrate CivBench's potential to estimate strategic capabilities in Civilization V as an unsaturated benchmark, reveal model-specific effects of agentic setup, and outline distinct strategic profiles not visible through outcome-only evaluation.
\end{abstract}

\section{Introduction}

AI agents are increasingly asked to make decisions that unfold over long horizons and interact with other actors' choices, from operational planning to autonomous execution. While evaluating agentic systems' strategic decision-making capabilities is essential, proper evaluation requires environments that are \textbf{generative} (procedurally created and continually evolving under imperfect information), \textbf{competitive} (multiple agents pursuing conflicting goals), and \textbf{longitudinal} (decisions compounding over hundreds of steps). As of now, few benchmarks provide all three.

LLM benchmarks increasingly adopt interactive environments \citep{Dynabench,AgentBench,WebArena,OSWorld} to overcome contamination and saturation issues for static benchmarks \citep{InvestigatingDataContamination,OpenSourceDataContaminationReport,Dynabench,SystematicStudyofBenchmarkSaturation}, yet limitations persist: many evaluate task completion in isolation, and game-based benchmarks with direct competition often resort to simplified settings or mini-games \citep{CivRealm}, constraining evaluation to step-level reasoning rather than long-horizon optimization \citep{DeepPlanning}. Evaluation methodology also lags: many benchmarks collapse each episode to a terminal signal or rely on hand-crafted checklists \citep{AgentBoard,AgentQuest}. The issue is most acute in long-horizon multiplayer games, where win/loss signals are exceptionally sparse.

This paper introduces CivBench, a benchmark for evaluating LLM strategists in multiplayer Civilization V, a turn-based 4X strategy game (eXplore, eXpand, eXploit, eXterminate). Each player in the game leads a civilization on a procedurally generated map for up to 500 turns, making decisions across economy, technology, policies, military, and diplomacy. Players win by being first to satisfy one of several victory conditions: Domination, by capturing all opposing original capitals; Science, by completing a late-game space project; Diplomacy, by securing enough World Congress/United Nations votes; Culture, by becoming culturally influential over all rivals; or Time, by having the highest score when the turn limit is reached. \textbf{An LLM strategist denotes a model-condition pair}: a main LLM under a specific agentic setup. Each turn, the strategist reads a structured game-state report and sets macro-level controls, while the game's built-in AI executes tactics (Figure~\ref{fig:interface-example}). As such, CivBench measures strategic decision-making as expressed through this interface and a fixed executor in Civilization V (Section~\ref{sec:limitations}).

To address the measurement gap, this paper brings a progress-based estimation approach from RL/sport analytics into LLM benchmarking: train ML models on turn-level game state to estimate a \textbf{state-value function}, the probability that a player eventually wins the game given its current state, then transform the estimates into within-game competitive standing for evaluation purposes. We validate it through a triple-validity framework: predictive (accuracy in outcome prediction), construct (interpretable, strategically plausible features), and convergent (consistency across estimators). Combined with turn-level decision trajectories, our approach enables strategic profiling, characterizing how LLM strategists pursue victory.

We conducted an empirical study around three research questions within the game: (1) Can our progress-based estimation approach reliably produce turn-level signals for evaluating strategic capability? (2) What differences in strategic capability can CivBench reveal about LLM strategists? (3) What differences in strategic profiles can CivBench reveal about LLM strategists?

The paper makes two main contributions. First, we present an evaluation framework for LLM strategists in a generative, competitive, and longitudinal environment, paired with a progress-based estimation methodology that produces dense signals from gameplay trajectories. Second, we provide an empirical validation and demonstration of the approach, revealing differences in estimated strategic capability and strategic profiles across LLM strategists that could be invisible to outcome-only evaluation.

\section{Related Work}
\subsection{From Static Evaluation to Generative, Competitive, and Longitudinal Evaluation}
Static LLM benchmarks are increasingly vulnerable to data contamination and overfitting \citep{InvestigatingDataContamination,OpenSourceDataContaminationReport,Dynabench,StatictoDynamicEvaluationSurvey,WhenBenchmarksareTargets}, leading to the rise of interactive environments for dynamic evaluation. Agents act and observe changes over longer horizons, exposing weaknesses less visible in static benchmarks \citep{AgentBench,SOTOPIA,SurveyonEvaluationofLLMbasedAgents,EvaluationandBenchmarkingofLLMAgentsSurvey}, e.g., in open-ended computer use and consequential task execution \citep{WebArena,OSWorld,WorkBench}. However, many interactive benchmarks evaluate task execution in isolation, missing multi-agent competition, and emphasize local, step-level reasoning over ``true'' planning ability \citep{DeepPlanning}.

Combining measurable objectives, dynamic opponents, and longitudinal validation, multiplayer strategic games (e.g., Diplomacy or Civilization) are rich playgrounds for evaluating strategic decision-making. Yet, few game-based benchmarks fully leverage this potential: some reduce strategic reasoning to formal game-theoretic computations or narrow negotiation protocols \citep{TMGBench,BattleAgentBench,DigitalPlayer}; some distribute evaluation across suites of ``mini'' games without sustained competition \citep{LLMArena,GameBench,BALROG,SPINBench}; others embed agents in deeper strategic environments but have an RL core \citep{DiplomacyCicero}, restrict play to one-on-one matchups \citep{TextStarCraftII}, or center on human--AI coordination \citep{HIVE}. Multiplayer strategy environments can also reveal behavioral tendencies. In scripted military and diplomatic scenarios, LLMs exhibit escalation risks, including first use of nuclear weapons \citep{rivera2024escalation,lamparth2024human,xu2025nuclear,payne2026aiarms}. On the other hand, while the depth of generated strategy games brings exciting opportunities to understand models' emergent tendencies, it also brings on a novel challenge. For example, operating on the tactical level, CivRealm found both RL and LLM agents struggle to make substantial progress in full Civilization games, leaving assessment largely out of reach \citep{CivRealm}.

\subsection{The Need for Progress-based Evaluation for Agentic Systems}
Another challenge lies in evaluation methodology, where terminal outcomes alone are too sparse for effective evaluation of multiplayer settings. Win/loss outcomes provide a coarse view of performance that cannot capture behavioral diversity \citep{StarCraftIIBenchmarkforAccessibleReinforcementLearning}, underestimating agents' planning quality, trajectory stability, and recovery over time \citep{VendingBench,DeepPlanning,EcoGym}. Some models can articulate optimal strategies but fail to execute them, creating an invisible knowing-doing gap \citep{BALROG}. The information loss compounds in multiplayer settings, where even long-horizon games produce one signal per game.

The sparsity of ternary reward signals in competitive games has long motivated intermediate measurement: in-game win-probability models are standard diagnostics in sports analytics \citep{SoccerWinProb}, and learned value functions over game-state features served as dense progress signals for reinforcement learning work, such as AlphaStar \citep{AlphaStar} and OpenAI Five \citep{OpenAIFive}.

While recent benchmarks start to introduce progress-based metrics, most still rely on hand-crafted checkpoints \citep{AgentBoard,AgentQuest,MultiAgentBench}. In LLM training, process reward models show how dense, step-level signals carry information that terminal outcomes alone miss \citep{LetsVerifyStepByStep}. However, existing work, such as SPINBench's move-accuracy analysis, focuses more on error rate than in-progress diagnostics \citep{SPINBench}.

\section{CivBench Design}
\subsection{Vox Deorum: Benchmark Environment}
CivBench is built on \textbf{Vox Deorum} \citep{VoxDeorum}, an open-source infrastructure that communicates with the Civilization V game with an agentic infrastructure for harness and telemetry. As a 4X grand strategy game, Civilization V requires players to make coordinated decisions across economic development, military strategy, diplomatic relations, and technological advancement over hundreds of turns (longitudinal), under ever-evolving situations (generative), imperfect information, and multilateral competition (competitive). Beyond AI research, Civilization has also served as a proxy environment for human decision-making research in management education \citep{Simons2021GoodGamers}.

Vox Deorum is built upon the Vox Populi (community patch) mod, which has substantially improved the base game's AI. It operationalizes macro-level strategic control by replacing Vox Populi AI (VPAI)'s strategic module with LLM strategists, delegating tactical execution back to the in-game AI (see Appendix~\ref{sec:appendix-agent-configurations} for details). 

With Vox Deorum, the LLM strategist each turn receives a structured Markdown report summarizing victory progress, available strategic options, player and city states, military conditions (including limited analysis from VPAI), and recent events. The strategist directly controls victory target (domination, science, culture, and diplomacy), technology research, and social policy adoption (simulated as persistent modifiers in the game). It also shapes tactical decisions through 34 \textbf{flavors}, numeric weights (0 to 100) that weigh on VPAI's tactical algorithms, such as offense (weight for tactical AI's offensive posture and desire to sacrifice units), expansion (weight for settling new cities), or gold (weight for prioritizing gold income in building and tile selection). It influences diplomacy through relationship modifiers (-100 to 100) to every other player and 26 diplomatic persona values (0 to 10). Figure~\ref{fig:interface-example} shows an abridged strategist turn from a benchmark game.

\begin{figure}[t]
\centering
\scriptsize
\setlength{\fboxsep}{4pt}
\fbox{\begin{minipage}[t]{0.545\linewidth}
\textbf{Turn input report (abridged)}\\[2pt]
\texttt{\# Situation}\\
MaxTurns: 500; MapSize: Standard; ``You are Harun al-Rashid, leader of Arabia (Player 3)''\\[2pt]
\texttt{\# Options}\\
GrandStrategies: Culture, UnitedNations, Spaceship, Conquest; Flavors (34): Offense, Defense, Expansion, Science, \ldots\\[2pt]
\texttt{\# Victory Progress}\\
Cultural: Arabia influential over 3/4 rivals; The Inca: Popular (99\%), 2 turns to Influential. Domination: The Inca control 3/8 capitals \ldots\\[2pt]
\texttt{\# Players}\\
Player 0 (The Inca): Score 5722, Era Information, Cities 22, \ldots{} TheirOpinionOfUs: ``The Inca are jealous of our superior technology!'' \ldots\\[2pt]
\texttt{\# Briefings} (Briefed condition only)\\
Military: ``Tenochtitlan \ldots{} Friendly Strength $\approx$2{,}700; Enemy $\approx$6{,}200; Posture: Hedgehog \ldots''
\end{minipage}}
\hfill
\fbox{\begin{minipage}[t]{0.39\linewidth}
\textbf{Strategist decisions (tool calls)}\\[2pt]
\texttt{set-flavors}\\
GrandStrategy: Culture; Science: 100; Culture: 40; Gold: 40\\
Rationale: ``With The Inca becoming Influential in 2 turns, our cultural influence over all civilizations will be complete. The new bottleneck is unlocking the Citizen Earth Protocol wonder \ldots{} pivoting to maximum science urgency \ldots''\\[3pt]
\texttt{set-research}\\
Technology: Alternative Energy\\
Rationale: ``\ldots{} Citizen Earth Protocol is gated behind Future Tech, reached via Robotics. Alternative Energy directly unlocks Robotics \ldots''\\[3pt]
VPAI then executes tactics (unit movement, city management, and combat) under the updated weights.
\end{minipage}}
\caption{An example, abridged strategist turn from a game. Appendix~\ref{sec:appendix-agent-configurations} documents the full interface.}
\label{fig:interface-example}
\end{figure}

\subsection{Flexible Setups for an Ever-Evolving Agentic Landscape}
CivBench is designed with the ever-evolving agentic landscape in mind: since we leverage statistical models for estimating multiplayer strength, benchmark runs can mix multiple configurations, up to 42 players per game. Civilization assignment should be randomized to balance per-civilization effects, and VPAI is recommended as a baseline check.

Each benchmark run should be conducted under consistent game settings: ruleset (map script, map size, player number, game speed, victory condition, handicap, etc.) and starting condition (e.g., from turn 0 or from existing savegames). This study used defaults throughout to keep the schedule tractable.

\subsection{Estimating LLM Strategists' Capability Through A Progress-based Approach}
We define strategic capability as a strategist's ability to convert decisions into winning potential in long-horizon, multiplayer Civilization V play. Augmenting sparse win/loss signals, CivBench trains ML models to estimate a \textbf{state-value function}: the probability that a player eventually wins the game, conditional on the turn-level game state. We use these turn-level estimates as measurement proxies for within-game competitive standing, and evaluate candidate estimators along three dimensions: \textbf{predictive validity}, \textbf{construct validity}, and \textbf{convergent validity} (see Section~\ref{validity}).

For \textbf{per-game estimation}, we aggregate turn-level predicted probabilities into a measure of within-game competitive standing. We compute each player's progress-weighted (0-1) average of turn-level estimates, then normalize to \emph{relative standing} against the strongest player in that game. Since the eventual winner can differ from the strongest player throughout the game (e.g., a come-from-behind Science victory), we apply a winner-preserving consistency correction when summarizing within-game standing (146/551 games in the full estimation corpus; 26.5\%). Removing this step leaves overall ELO ordering unchanged.\footnote{Rankings \#1--\#10 are identical; between \#11--\#15, some adjacent pairs flip, all within overlapping confidence intervals.} We then use OLS regression on the logit-transformed relative standing to remove civilization effects, yielding what we term \textit{revised standing}.\footnote{Civilization assignment is randomized but not balanced across conditions. Without the adjustment, strategists stay in their tiers but may shift 2--3 positions.}

Across games, we estimate each strategist's underlying capability as overall \emph{worth} (ELO) via Bradley-Terry models fitted on within-game revised standings. Because agents in this study lack cross-game memory, we assume stationary capabilities throughout the benchmark. To understand how many games are necessary to provide sufficient evidence for stable ratings, we conducted a per-player-type convergence ablation (Appendix~\ref{sec:appendix-bt-convergence}).

\subsection{Mapping LLM Strategists' Strategic Profiles through Gameplay Trajectories}

We characterize each LLM's strategic behavior through turn-level decision-making trajectories: per-turn sequences of logged game-state signals paired with the strategist's decisions and written rationales (see Appendix~\ref{sec:appendix-logged-signals}). Game-state signals (e.g., population count, technology numbers, or military strength) are logged each turn from the observable game state. Decision and reasoning signals (e.g., policy choices, flavor changes, or LLMs' written rationale) are logged when a strategist makes a decision. We demonstrate three example analyses: \textbf{Strategy time allocation}, the mean fraction of game time each strategist spends targeting each of the four victory paths, tested against equal allocation (25\%) with t-tests. \textbf{Strategic commitment}, how concentrated the allocation is on the strategist's dominant victory pursuit relative to VPAI's baseline commitment level (Welch's t-test). \textbf{Strategy adaptation}, \emph{strategy pivots}, turns where the victory pursuit changes (excluding the first 25 turns to filter noise), reported as frequency per game, predicted win probability at pivot time, and transition flows.

\section{Empirical Study}
\subsection{Benchmarked 7 LLMs Across 12 Matchup Combinations}
\label{sec:experiment-setup}
We benchmarked 7 LLMs (Claude Sonnet-4.5, Kimi-K2.5, MiniMax-M2.5, Qwen-3.5, GLM-4.7, DeepSeek-V3.2, GPT-OSS-120B) across 12 matchup combinations, totaling 307 games and 1,674 LLM player-games. We ran 194 VPAI self-play games to supplement estimators' training data. Each game has 8 players, on the Communitas\_79a map script (which generates Earth-like planet maps and is the most popular among the community), and default rulesets. Due to budgetary constraints, our matchup schedule is unbalanced (Appendix Table~\ref{tab:appendix-matchup-schedule}), and Sonnet-4.5 has the fewest games (28 per configuration) with likely unsaturated ratings. We quantify the resulting uncertainty through confidence intervals, a convergence ablation (Appendix~\ref{sec:appendix-bt-convergence}), and a game-level bootstrap (Appendix~\ref{sec:appendix-bootstrap}).

We tested two CivBench strategist configurations (Appendix~\ref{sec:appendix-agent-configurations}) to probe how a fixed agentic setup interacts with strategist models: Simple, in which the agent receives the standard game-state report, and Briefed, in which detailed game-state sections (cities, military, events) are summarized by a briefing subagent into structured military, economy, and diplomacy briefings. The strategist still sees the game summary and can steer each briefer via natural-language instructions. To simulate cost-constrained deployment, a fixed weaker model (GPT-OSS-120B) generates all briefings.

This study included a \emph{null} agent, a partial ablation of VPAI that estimates a capability floor: each turn, it forces all strategic parameters to baseline and randomly selects technology and policy, leaving VPAI only its victory target and some diplomatic actions.

\subsection{Seven Machine Learning Models as Estimators}
We evaluated seven ML models as candidate estimators (see Appendix~\ref{sec:appendix-estimation-specs} for features, architectures, and hyperparameters). \textbf{Score} transforms each player's share of in-game score through a tuned exponent into a victory probability. It achieves reasonable accuracy but only for Domination victories, an issue anecdotally observed by Civilization players \citep{Civ5ScoreThread,Civ5AIGame4UCStats}. This motivates the inclusion of 23 game-state features (Appendix~\ref{sec:appendix-estimation-specs}) spanning technology, policy, economy, military, diplomacy, and culture, encoded as within-game shares and gaps. All learned models below take these features as input and train on terminal win/loss labels.

\textbf{Baseline} is a logistic regression over the 23 features, trained with cross-entropy and isotonic calibration. Its linear coefficients aid interpretation but cannot capture feature interactions. \textbf{XGBoost} \citep{XGBoost} is a gradient-boosted tree ensemble over the same features plus \emph{turn\_progress} (=\emph{turn} / \emph{max\_turn}), trained with calibrated logistic loss to learn nonlinear interactions. \textbf{MLP} \citep{MLPUniversalApprox} is a one-hidden-layer neural network over the same inputs, trained with progress-weighted cross-entropy. These three operate per player, missing between-player interactions and cross-player calibration. \textbf{GroupedMLP} scores every player in a game-turn with a shared network and applies a softmax across players, hence group-wise cross-entropy jointly calibrates each turn's probabilities. \textbf{InteractionMLP}, a DeepSets-like model \citep{DeepSets}, encodes all players' features together through a shared encoder and pooled context before decoding per-player utilities under the same group-softmax objective. \textbf{AttentionMLP} replaces fixed pooling with learned multi-head self-attention across players \citep{Transformer}.

All tunable models are optimized via Optuna \citep{Optuna} (1,000 trials each) with a train/validation gap penalty to suppress overfitting. Unless otherwise noted, all models were trained and evaluated with 5-fold cross-validation grouped by game, and all reported metrics are computed from out-of-fold predictions. As a robustness check on cross-domain generalization, we trained and evaluated the models under an LLM/non-LLM split.

\section{Benchmark Results}

\subsection{Evaluation of Progress-based Estimators}
\label{validity}
\textbf{Predictive validity} is tested through the held-out accuracy check, where all metrics come from out-of-fold predictions under five-fold cross-validation grouped by game. Appendix Table~\ref{tab:predictive-validity} reports metrics for all models. Loss declines monotonically with turn progress (Appendix Figure~\ref{fig:appendix-loss-by-progress}), reflecting the competitive nature of long-horizon strategic games. Score captures ranking signals (AUC 0.825), but the victory-type breakdown (Appendix Figure~\ref{fig:late-game-logloss-by-victory-type}) reveals poor non-Domination prediction. While the three MLP variants correlate closely, we chose AttentionMLP (with the smoothest log loss by victory type) as the primary estimator for downstream analyses.

\textbf{Construct validity} asks whether the estimators' learned signals triangulate with domain knowledge. Baseline's coefficients recover a plausible linear structure: \texttt{technologies\_gap} ($\beta$ = -0.69), \texttt{score\_ratio} ($\beta$ = +0.53), and \texttt{policies\_gap} ($\beta$ = -0.44) are the largest-magnitude mean coefficients, though some (e.g., negative \texttt{science\_share}) resist isolated interpretation due to collinearity. For AttentionMLP, the permutation analysis (Appendix~\ref{sec:appendix-estimation-specs}) shows that growth, score, and influence all matter, with some shifts by victory type. War-group features remain weak even for Domination, likely because in-game score already captures military advantage and snapshot-based architectures miss temporal conflict dynamics.

\textbf{Convergent validity} is measured by internal consistency. Trained on the same corpus, inter-model agreement is robust across all seven models and between game types. Within-game rank agreement is high overall and grows stronger in later game phases (Figure~\ref{fig:rank-agreement-by-progress}). Bradley-Terry ratings derived independently from each model yield consistent orderings on the full evaluation set (mean pairwise $\rho$ = 0.925). 

Additionally, we probed model generalization across data distributions. In the LLM/non-LLM split, which retrains the estimators on VPAI self-play games and evaluates them on LLM games, predictive accuracy remains close to the pooled setting (Appendix Table~\ref{tab:predictive-validity}), and the induced orderings still agree (mean pairwise $\rho$ = 0.916). Preliminary results from an upcoming, more controlled CivBench dataset, which reuses these trained estimators on 485 new games (including 24 VPAI and 24 Null self-play games), suggest predictive accuracy transfers: AttentionMLP reaches AUC 0.860, Brier 0.080, roughly equal to the numbers from this paper's dataset.

\begin{figure}[hbt!]
\centering
\includegraphics[width=0.82\linewidth]{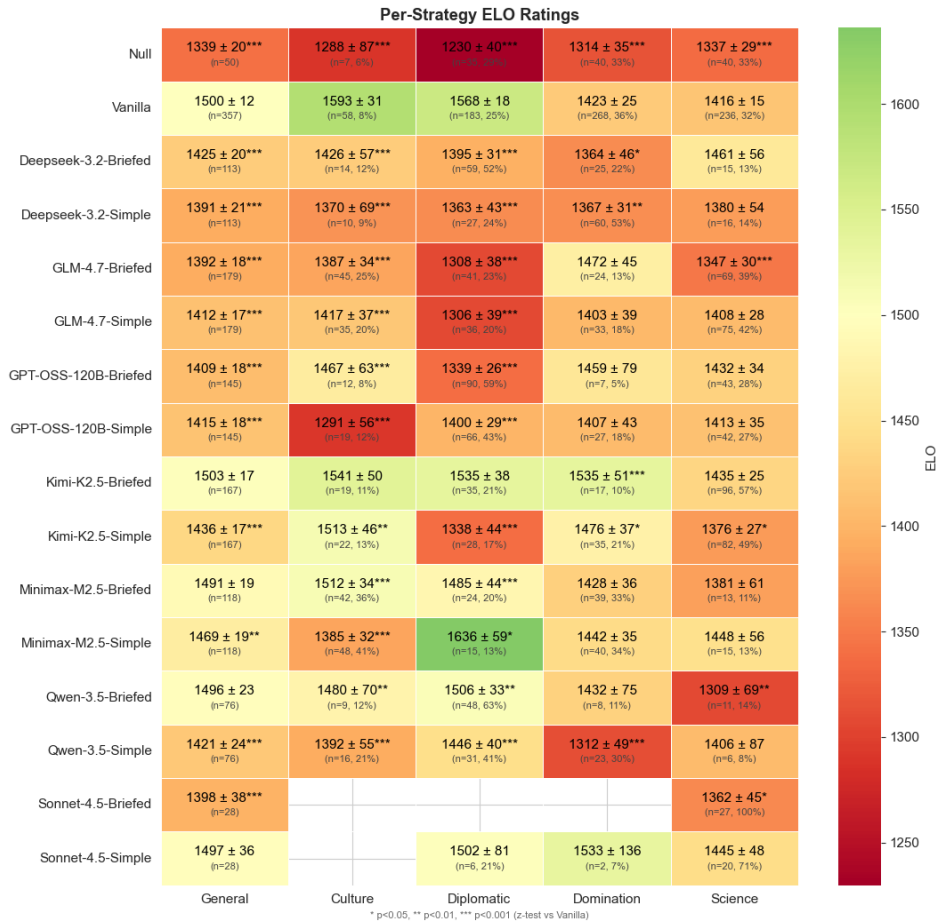}
\caption{Overall (first column) and per-strategy (rest columns) ELO ratings.}
\label{fig:overall-and-strategy-elo}
\end{figure}

\subsection{Estimating LLM Strategists' Strategic Capability}
Several LLM strategists approach VPAI's rule-based strategic module, yet none can consistently exceed it. Four strategists show no significant difference from VPAI (1500): Kimi-K2.5-Briefed (1503, p = 0.6947), Sonnet-4.5-Simple (1497, p = 0.8811), Qwen-3.5-Briefed (1496, p = 0.7367), and Minimax-M2.5-Briefed (1491, p = 0.3635). All LLM strategists outperform the Null ablation (1339), where VPAI's strategic module is mostly suppressed. The Briefed configuration produces model-specific tradeoffs. Kimi-K2.5 gains +67 ELO (1503 vs 1436), Qwen-3.5 gains +75 (1496 vs 1421), but the effect is not uniform: Sonnet-4.5 drops 99 points under Briefed (1398 vs 1497). Given the unbalanced schedule, rankings carry uncertainty: the game-level bootstrap (Appendix~\ref{sec:appendix-bootstrap}) shows adjacent orderings remain uncertain, while broad tier separations are stable.

Estimated capability can vary dramatically across strategists' dominant victory pursuit, exposing strategists' distinct strengths and weaknesses (Figure~\ref{fig:overall-and-strategy-elo}). Some strategists may exceed VPAI in certain victory types: for example, Minimax-M2.5-Simple on Diplomatic (ELO 1713, p < 0.001, n = 15) and Kimi-K2.5-Briefed on Domination (ELO 1612, p < 0.001, n = 17). The Briefed condition generally improves culture victory capability (except GLM-4.7) but shows mixed effects elsewhere.

\subsection{LLMs' Strategic Profiles}
\label{sec:strategic-profiles}
\textbf{Strategy preferences.} Strategists differ in victory-target time allocation (Appendix Figure~\ref{fig:strategy-time-allocation}). Sonnet-4.5-Briefed and Sonnet-4.5-Simple are the most extreme, dedicating 77.6\% and 65.0\% of game time targeting science victory. GPT-OSS-120B-Briefed is the most diplomatic-oriented LLM at 51.7\%, while Minimax-M2.5 strategists stand out for heavy Culture allocation (35.6\%-41.8\%) alongside substantial Domination allocation (32.7\%-33.0\%). The briefed configuration shifts preferences in model-specific ways: e.g., Sonnet-4.5 becomes more science-focused under briefing (from 65.0\% to 77.6\%), while Deepseek-3.2 shifts toward diplomatic (from 27.4\% to 46.7\%).

\textbf{Strategic commitment.} Many models commit more to their chosen victory target than VPAI, though the pattern is not universal (Appendix Figure~\ref{fig:appendix-strategic-commitment}). Minimax-M2.5 and Sonnet-4.5 strategists are especially committed: Minimax-M2.5 allocates far more time to its dominant path than VPAI, and Sonnet-4.5 hyper-commits to science (+17.3 to +21.5 points versus Vanilla).

\begin{figure}[hbt]
\centering
\includegraphics[width=0.82\linewidth]{notebooks/extracted/profiles/victory\_commitment/images/cell\_17\_out\_1.png}
\caption{Most chosen strategy, most pivoted-to strategy, and best-ELO strategy.}
\label{fig:chosen-pivoted-best-elo}
\end{figure}

\textbf{Strategy adaptation.} Models differ in how frequently and when to pivot victory pursuit (Appendix Figure~\ref{fig:win-probability-at-pivot}). Most LLMs fall between roughly 2 and 6 pivots/game, while Vanilla is far more active at about 19.6 pivots/game. Most pivots occur at low win probability, showing a primarily reactive pattern. Appendix Figure~\ref{fig:appendix-pivot-flow-heatmaps} shows the diverse pivot pattern across models. Each model's pivot flows remain similar across briefed and simple configurations (mean r = 0.76 vs. 0.07 for cross-model pairs). Notably, this analysis omits secondary pathways, which is sometimes described by LLMs without direct support from the interface.

\section{Discussion}

\subsection{Progress-Based Evaluation: Validation and Generalizability}
CivBench's progress-based approach constructs a \textbf{dense, behaviorally anchored proxy} from gameplay trajectories, capable of enhancing terminal win/loss signals to estimate within-game competitive standing with far fewer games.

Benchmarks must validate their \emph{evaluation signals}. Civilization's in-game score seemed like a natural proxy, yet we found it strongly biased towards Domination victory. The loss progression from Score to regression to MLP models shows the impact of capturing between-player interactions, and even the attention-based model is likely unsaturated: war-related signals are temporal (e.g., campaign momentum), which single-turn estimators struggle to capture.

We believe more benchmarks should apply CivBench's progress-based approach, which transfers readily to domains with turn-by-turn observable signals and well-defined terminal outcomes: train on turn-level state to predict terminal outcomes, aggregate into standing estimates, and validate through the triple-validity lenses.

As with human chess or Go ratings, CivBench was designed to support runs that are not fully crossed, including future human participation. Since decision-making performance is contingent on the pool of opponents, CivBench produces relative ratings, conditional on the sampled schedule and opponent pool. This design also carries a practical benefit, as a useful benchmark can estimate a newcomer's capability against a subset of opponents, not by rerunning the entire match. While agents can adapt within games, this study assumes \textbf{stationary abilities} across games. Should agents exhibit cross-game learning, play-order-sensitive rating systems such as OpenSkill \citep{OpenSkill} would be appropriate.

\subsection{Strategic Capability: An Unsaturated Benchmark}
\label{sec:discussion-benchmark}
The field needs benchmarks where agentic AI has \emph{not yet saturated} \citep{SystematicStudyofBenchmarkSaturation,WhenBenchmarksareTargets}, a space that CivBench targets. While Civilization V has extensive public guides and discussions that LLMs likely saw in training, domain knowledge on the game is different from data contamination on benchmark questions. Since each game is a procedurally-generated, dynamically-evolving complex system, each turn represents a unique situation that requires in-context adaptation rather than recalled strategies. Moreover, as a free-for-all game, players may scheme against early leaders, penalizing beelining strategies. Our example of Sonnet-4.5 illustrates this: both conditions hyper-commit to science in nearly 90\% of 28 games, and their overlapping victory targets only lower each other's chances. Thus, CivBench remains a wide-open challenge: while all LLM strategists exceed the Null lower bound, only four roughly match VPAI's strategic module. This achievement alone is nontrivial, as LLM strategists receive minimal prompting, while VPAI is an expert-crafted rule-based system with more than a decade of open-source contributions.

As modern benchmarks start to target entire agentic systems rather than foundational models alone \citep{li2026liveagentbench}, CivBench measures LLM strategists, i.e., the \textbf{full agent setup} (foundational model, agentic setup, and interface with the VPAI tactical executor). Like written examinations, which test both subject knowledge and pen familiarity, agentic benchmarks measure situational performance rather than absolute latent capacities.

Delegating tactical execution creates both benefits and limitations. On the downside, the interface between LLMs and VPAI's tactical layer creates a confounding factor. CivBench's isolation of strategic actions (e.g., resource prioritization or diplomatic postures) is imperfect, and the flavor weight interface can be opaque (e.g., what is the meaning of offense = 75). As such, the captured signals reflect both strategic decision quality and the ability to operate and adapt through the interface. The Null ablation partially bounds this confound: it keeps the executor while suppressing strategic control, and every LLM strategist improves over its 1339 ELO, which suggests strategic input contributes variance beyond the executor alone.

On the up side, this interface appears to elicit LLM strategists' exploratory behavior in our study, i.e., applying incremental changes and observing tactical impacts before moving forward. Since LLMs struggle with tactical-level decisions alone \citep{CivRealm, HIVE}, leading to knowledge-doing gaps \citep{BALROG}, the delegation also helps measure their strategic decision-making without getting lost in tactical abilities. That said, whether our findings would hold if agents executed low-level actions directly remains an open empirical question. Based on CivRealm's results, we expect performance to degrade substantially before strategic differences could be observed. Future work may instead decouple language-level strategic reasoning from interface operation, for example by having the strategist reason purely in language before a separate executor converts the strategy into tool calls \citep{szot2026exploration}, though any such pipeline adds its own confound.

Our study identifies model-specific interaction effects between LLMs and agentic "briefer" setups (Kimi +67, Qwen +75, Sonnet -99), highlighting the need for evaluating combinations of models and agentic setups. More intervention-heavy strategists (Qwen-3.5, Kimi-K2.5, Deepseek-3.2), measured by briefer instruction counts (Appendix Figure~\ref{fig:appendix-briefer-instruction-intensity}), tend to be among the stronger beneficiaries of briefing, though outliers remain: Sonnet-4.5 intervenes frequently yet degrades, while Minimax-M2.5 changes little despite relatively light instruction. A deeper investigation of briefer-strategist interactions warrants a further study.

\subsection{Strategic Profiles: CivBench as a Diagnostic Tool}
Moving beyond capability alone, CivBench provides a decision-making \emph{contour} to explore and diagnose LLM strategists. This study demonstrates three facets: victory preferences, commitment intensity, and adaptation. Foundational models shape LLM strategists' victory preferences, which the Briefed condition can shift in model-specific ways (Section~\ref{sec:strategic-profiles}). 

Combining progress-based estimation with strategic pivots reveals strategy-level differences invisible in aggregate rankings. For example, preferences and strength do not always align (Appendix Figure~\ref{fig:chosen-pivoted-best-elo}): Sonnet-4.5-Simple consistently pursues Science victory, yet its real strength may lie elsewhere. MiniMax-M2.5-Simple predominantly chooses and pivots to Culture, yet achieves its best ELO in Diplomatic (1659.7). Moreover, most LLM strategists pivot victory targets under distress: compared with VPAI's strategic module, most pivots occur well below the expected win probability in 8-player games (0.125). 

Is the misalignment between preferences and strengths a weakness or a natural response to competitive pressure? Is reactive pivoting a sign of limited strategic initiative, or of persistence beyond rule-based systems? While CivBench detects these patterns, deeper questions remain for future studies: savegame replay (loading identical game states) can enable controlled comparison of strategic responses, and flavor-level analysis triangulated with reasoning trajectories can better detect the knowing-doing gap.

\section{Limitations}
\label{sec:limitations}
In addition to those discussed above, we also recognize the following limitations:

\textbf{Transfer beyond Civilization V.} CivBench measures strategic decision-making within Civilization V, and we make no claim that its scores proxy real-world strategic competence. As a descriptive sanity check (n = 7), base models' mean CivBench ELO correlates modestly and non-significantly with Arena (Spearman's $\rho$ = 0.393, p = 0.383) and GPQA Diamond ($\rho$ = 0.429, p = 0.337), and strongly with SWE-bench Verified ($\rho$ = 0.929, p = 0.003).

\textbf{Multi-skill confound and interface mediation.} CivBench performance reflects a bundle of skills necessary for demonstrating strategic decision-making capability within the game: strategic reasoning, Civilization-domain knowledge, interface use, long-context retrieval, and spatial abstraction from structured state. Since all measured behavior is mediated by the fixed VPAI executor and flavor weight interface (Section~\ref{sec:discussion-benchmark}), CivBench cannot, and does not intend to, cleanly isolate pure strategic reasoning capabilities.

\textbf{Single map script and schedule imbalance.} All games use the Communitas\_79a map script with randomized maps, and generalization across map scripts remains untested. The matchup schedule is budget-constrained and unbalanced, with Sonnet-4.5 having the fewest games. While the convergence ablation (Appendix~\ref{sec:appendix-bt-convergence}) indicates broad-tier ratings stabilize within roughly 20 to 30 games for some player types and 50 to 60 for others under this distribution and the game-level bootstrap (Appendix~\ref{sec:appendix-bootstrap}) quantify the resulting uncertainty, nearby orderings should not be over-read.

\textbf{No human baseline.} While Vox Deorum supports interactive human play, a fair baseline would require hundreds of human playthroughs beyond this study's resources. Our upcoming controlled version of CivBench will make a limited baseline more feasible.

\label{sec:limitations-end}%
\section*{Acknowledgments}
This work used the Jetstream2 cloud environment at Indiana University through allocation CIS250934 from the Advanced Cyberinfrastructure Coordination Ecosystem: Services \& Support (ACCESS) program, which is supported by U.S. National Science Foundation grants \#2138259, \#2138286, \#2138307, \#2137603, and \#2138296. In addition, we acknowledge the support from Synthetic.new, National Research Platform (NRP), and Jerry Jiang (Akatos) for LLM inference support.

\section*{Ethics Statement}
This study involves no human subjects. All gameplay data were generated by LLM agents and rule-based AI, so no IRB approval or informed consent was required.

Generative AI is the study target of this paper and is also used, with sufficient human supervision, in several aspects: 1) the development of the data extraction and analysis code; 2) the production of visualization; and 3) the writing of this paper. In all cases, human researchers confirm the detailed plan before implementation and revise (in the paper's case, completely rewrite) the generated artifacts. Generative AI is also extensively used to double-check or provide feedback on references, data analysis, visualization, and the paper itself. 

Civilization V embeds particular historical and cultural narratives (e.g., civilizations as monolithic entities, militaristic framings of history). These representations are inherited by the benchmark and should not be read as endorsing any geopolitical or cultural worldview.

Given the strategic capability levels we estimated and the behavioral tendencies we observed in LLM strategists (e.g., reactive pivoting under distress, hyper-commitment to single victory paths, or some models' hyper-focus on nuclear weapons), we do not recommend applying similar agentic systems to real-world decision-making domains at this point.

\section*{Reproducibility Statement}

We take several steps to support the reproducibility of both the benchmark environment and the empirical results reported in this paper.

\textbf{Benchmark system and code.} The Vox Deorum system \citep{VoxDeorum}, which provides the underlying technical infrastructure, is open-sourced at GitHub. CivBench's benchmark configurations, prompt configurations, evaluation model architectures and hyperparameters, and all training and validation scripts will be released at [URL withheld for review; included as supplementary material].

\textbf{Estimation models and statistical analysis.} All seven estimation models were trained with a fixed random seed (42), held constant throughout the study. Hyperparameter search configurations (Optuna \citep{Optuna}, 1,000 trials), cross-validation splits (5-fold grouped by game), and trained model weights are included in the release. All Jupyter notebooks producing the paper's figures and tables are included with pre-computed cell outputs, allowing inspection and verification without rerunning. To verify that reported ratings are stable given the sample size, we conducted a Bradley-Terry convergence ablation (Appendix~\ref{sec:appendix-bt-convergence}), showing that some player types stabilize by roughly 20-30 games while others remain volatile until roughly 50-60 games.

\textbf{Data release.} Turn-level and game-level statistics (about 777 MB) sufficient to reproduce all statistical workflows, figures, and tables in the paper will be publicized alongside the code. Raw game trajectories (SQLite databases) and agentic workflow telemetry (OpenTelemetry format), totaling about 644 GB, are available to the academic community upon request.

\textbf{Experiment reproduction.} The matchup schedule is documented in Appendix Table~\ref{tab:appendix-matchup-schedule}. Game settings (map script, map size, game speed, victory conditions, handicap) are specified in Section~\ref{sec:experiment-setup} and the released configuration files. Exact game-level reproduction is infeasible due to Civilization V's inherent stochasticity; all claims are based on aggregate patterns across the full game corpus. In addition, we adopted fixed seating orders in experiment configurations. Analysis of VPAI self-play data shows no significant seating-order effect, but future experiments should still consider randomizing seating. 

\textbf{Compute requirements and costs.} Full reproduction of the study requires access to LLM inference services. Except for Sonnet-4.5 from Anthropic, all models are open-weight. The inference cost is estimated at \$10{,}497 before input cache or batch inference discount (Appendix~\ref{sec:appendix-llm-inference-costs}). The Civilization V instances were hosted at Jetstream2 with 4vCPU + 15GB CPU-only instances, although 8GB memory should suffice. Estimated CPU cost is $24$ hours/game $\times$ $4$ cores $\times$ $(307 + 50 + 194)$ games $= 52{,}896$ core-hours. The convergence ablation (Appendix~\ref{sec:appendix-bt-convergence}) suggests that a reduced reproduction could still produce meaningful tiers with fewer games for some player types.

\textbf{Practical access.} Running CivBench requires a licensed copy of Sid Meier's Civilization V with the Vox Populi mod. The Steam-hosted End User License Agreement for Civilization V\footnote{\url{https://store.steampowered.com/eula/8930_eula_0}} grants personal, non-commercial gameplay use and restricts commercial exploitation, redistribution, and modification except where separately authorized or documented. Users of CivBench remain responsible for complying with the applicable Steam, Take-Two, Civilization V, Vox Populi, and institutional terms.

\bibliography{civbench_refs}
\bibliographystyle{colm2026_conference}

\appendix
\section{Appendix / Supplementary Material}

\subsection{Appendix Figures and Tables}

\subsubsection{Matchup Schedule}

While we attempted to balance each strategist's appearance in the study, our matchup schedule was constrained by the study budget and availability of inference services.

\begin{table}[htbp]
\centering
\caption{Distinct matchup combinations used in the benchmark schedule. All listed LLMs appear in both Simple and Briefed configurations. The benchmark contains 12 distinct LLM matchup combinations, plus a separate Null ablation matchup against VPAI. Game counts are aggregated across runs with the same normalized matchup composition.}
\label{tab:appendix-matchup-schedule}
\small
\begin{adjustbox}{max width=\linewidth}
\begin{tabular}{l r r r}
\toprule
Matchup composition & VPAI seats & Games & LLM player-games \\
\midrule
Deepseek-3.2, GLM-4.7, Kimi-K2.5 & 2 & 93 & 558 \\
Deepseek-3.2, GPT-OSS-120B, Kimi-K2.5 & 2 & 2 & 12 \\
Deepseek-3.2, Kimi-K2.5, Minimax-M2.5 & 2 & 18 & 108 \\
GLM-4.7, GPT-OSS-120B & 4 & 66 & 264 \\
GLM-4.7, GPT-OSS-120B, Sonnet-4.5 & 2 & 10 & 60 \\
GLM-4.7, Kimi-K2.5, Minimax-M2.5 & 2 & 6 & 36 \\
GLM-4.7, Sonnet-4.5 & 4 & 4 & 16 \\
GPT-OSS-120B, Kimi-K2.5, Minimax-M2.5 & 2 & 7 & 42 \\
GPT-OSS-120B, Minimax-M2.5 & 2 & 11 & 66 \\
GPT-OSS-120B, Minimax-M2.5, Qwen-3.5 & 2 & 35 & 210 \\
GPT-OSS-120B, Sonnet-4.5 & 4 & 14 & 56 \\
Kimi-K2.5, Minimax-M2.5, Qwen-3.5 & 2 & 41 & 246 \\
Null & 4 & 50 & 200 (Null) \\
\bottomrule
\end{tabular}
\end{adjustbox}
\end{table}

\subsubsection{Evaluation Model Diagnostics}

\begin{table}[htbp]
\centering
\caption{Predictive validity metrics on the 5-fold CV vs. the LLM/non-LLM split.}
\label{tab:predictive-validity}
\small
\begin{adjustbox}{max width=\linewidth}
\begin{tabular}{l r r r r r r r}
\toprule
Model & AUC & Log loss & Brier & Split AUC & Split log loss & Split Brier & Delta LL \\
\midrule
Naive & 0.500 & 0.377 & 0.109 & 0.500 & 0.377 & 0.109 & +0.000 \\
Score & 0.825 & 0.291 & 0.088 & 0.813 & 0.297 & 0.089 & +0.006 \\
Baseline & 0.841 & 0.280 & 0.085 & 0.828 & 0.290 & 0.087 & +0.009 \\
XGBoost & 0.856 & 0.270 & 0.082 & 0.840 & 0.282 & 0.086 & +0.012 \\
MLP & 0.858 & 0.269 & 0.082 & 0.843 & 0.282 & 0.086 & +0.013 \\
GroupedMLP & 0.863 & 0.263 & 0.080 & 0.846 & 0.281 & 0.085 & +0.018 \\
InteractionMLP & 0.865 & 0.260 & 0.079 & 0.840 & 0.284 & 0.086 & +0.024 \\
AttentionMLP & 0.865 & 0.260 & 0.079 & 0.844 & 0.278 & 0.085 & +0.018 \\
\bottomrule
\end{tabular}
\end{adjustbox}
\end{table}

\begin{figure}[H]
\centering
\includegraphics[width=0.90\linewidth]{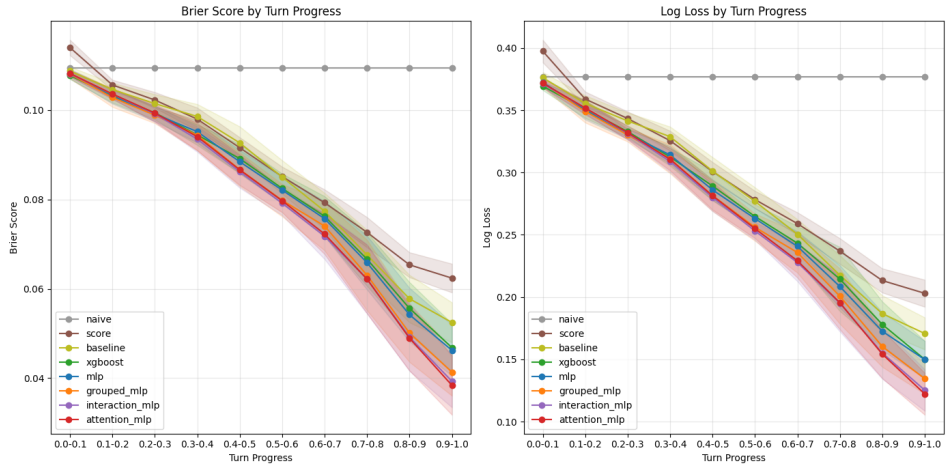}
\caption{Brier score and log loss by turn-progress decile for all eight models. All models improve as games resolve, with the learned models maintaining the lowest loss throughout.}
\label{fig:appendix-loss-by-progress}
\end{figure}

\begin{figure}[H]
\centering
\includegraphics[width=1\linewidth]{notebooks/extracted/models/visualize\_context\_slicing/images/cell\_08\_out\_0.png}
\caption{Late-game log loss by victory type and model.}
\label{fig:late-game-logloss-by-victory-type}
\end{figure}

\begin{figure}[H]
\centering
\includegraphics[width=1\linewidth]{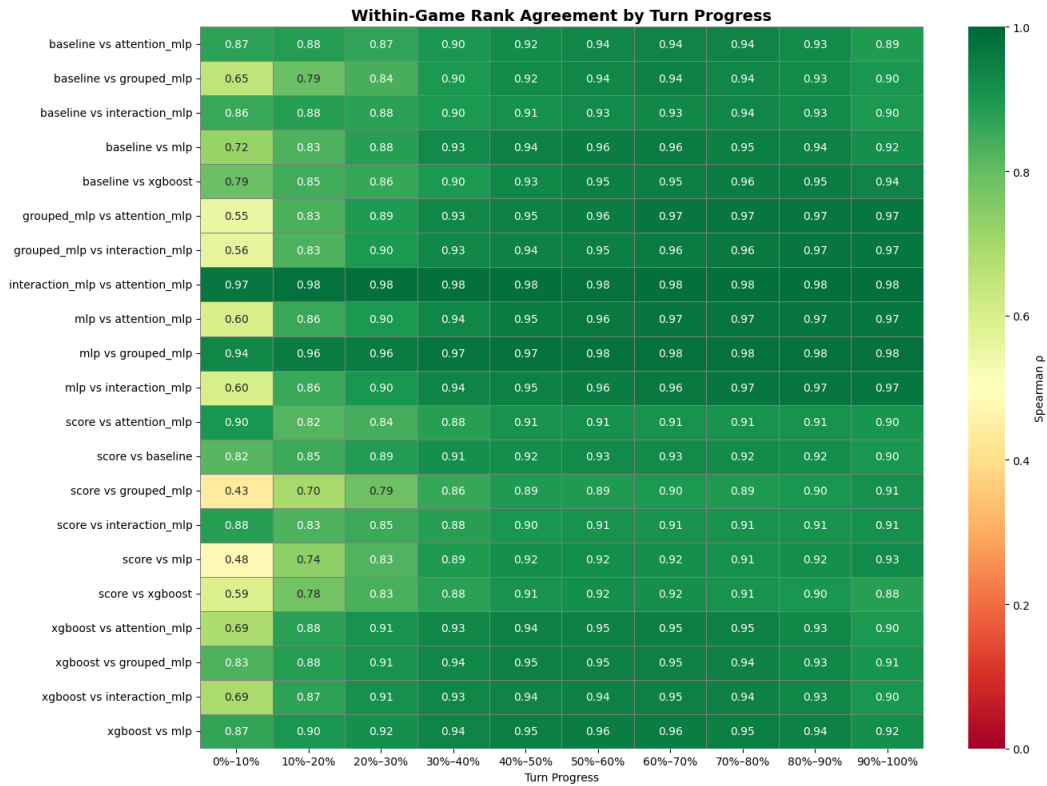}
\caption{Within-game rank agreement by turn progress. Agreement is high overall and grows stronger in later phases of the game.}
\label{fig:rank-agreement-by-progress}
\end{figure}

\begin{table}[H]
\centering
\caption{Top Baseline coefficients with approximate 95\% confidence intervals.}
\label{tab:appendix-baseline-coefficients}
\small
\begin{adjustbox}{max width=0.45\linewidth}
\begin{tabular}{l r r}
\toprule
Feature & Beta & Approx. 95\% CI \\
\midrule
\texttt{technologies\_gap} & -0.693 & [-0.846, -0.540] \\
\texttt{score\_ratio} & +0.525 & [+0.405, +0.645] \\
\texttt{policies\_gap} & -0.443 & [-0.602, -0.283] \\
\texttt{cities\_share} & +0.313 & [+0.126, +0.500] \\
\texttt{votes\_share} & +0.257 & [+0.190, +0.324] \\
\texttt{active\_wars} & -0.172 & [-0.223, -0.120] \\
\texttt{science\_share} & -0.170 & [-0.261, -0.078] \\
\texttt{military\_share} & +0.152 & [+0.069, +0.234] \\
\texttt{minor\_allies\_share} & +0.146 & [+0.094, +0.198] \\
\texttt{production\_share} & +0.134 & [+0.072, +0.196] \\
\bottomrule
\end{tabular}
\end{adjustbox}
\end{table}

\subsubsection{AttentionMLP Group Permutation Importance}

To complement the Baseline coefficient table, we probed the selected \textbf{AttentionMLP} with a post hoc group permutation analysis. The analysis is run on the fitted AttentionMLP without retraining. We group the model's inputs into the same nine semantic groups defined in the feature appendix (science, culture, economy, growth, religion, influence, war, welfare, and score), then permute all features in one group together.

Permutation is performed at the \texttt{(game\_id, turn)} level across players. For each repeat, a turn-level group receives another turn-level group's values for the chosen feature block, preserving within-group correlations among that block's features. Player positions are shuffled within the donor group to break the link between that semantic group and the true winner while retaining plausible cross-player structure. Importance is measured as the increase in log loss relative to the unpermuted model, aggregated here by eventual victory type over 30 repeats. We report group-level reliance rather than local, per-feature attributions.

This analysis is descriptive rather than causal. Low permutation importance can reflect redundancy with stronger features rather than irrelevance, and a grouped per-turn estimator can understate signals that are expressed mainly through trajectories rather than isolated snapshots.

\begin{figure}[H]
\centering
\includegraphics[width=0.72\linewidth]{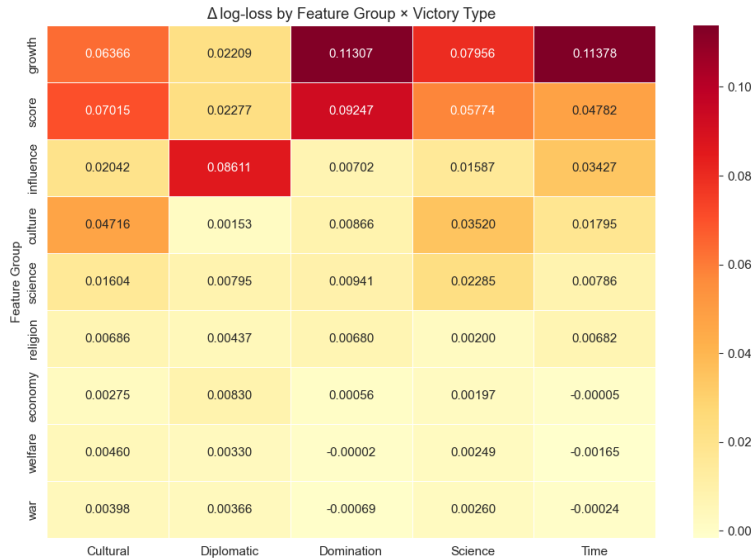}
\caption{Group permutation importance for AttentionMLP by feature group and eventual victory type. Each cell shows the increase in log loss after permuting one semantic feature group; higher values indicate that AttentionMLP relies more heavily on that group for that victory type.}
\label{fig:appendix-group-permutation}
\end{figure}

The victory-type breakdown clarifies two properties of the selected estimator. First, the model is not using score alone: growth and influence remain highly important, and their relative importance shifts by victory type. Second, the war group appears weak even for Domination. We do not interpret this as meaning war is strategically unimportant. Rather, in the current snapshot-based setting, much of Domination's win signal is already recoverable from \texttt{score\_ratio}, while the war group mostly captures transient conflict state (\texttt{military\_adj}, \texttt{military\_utilization}, \texttt{active\_wars}, \texttt{truces}) that is less informative one turn at a time than longer-horizon military momentum or campaign progression. Future work should explore sequence-aware or time-series estimators for strategic conflict dynamics.

\subsubsection{Strategic Capability Diagnostics}

\begin{figure}[H]
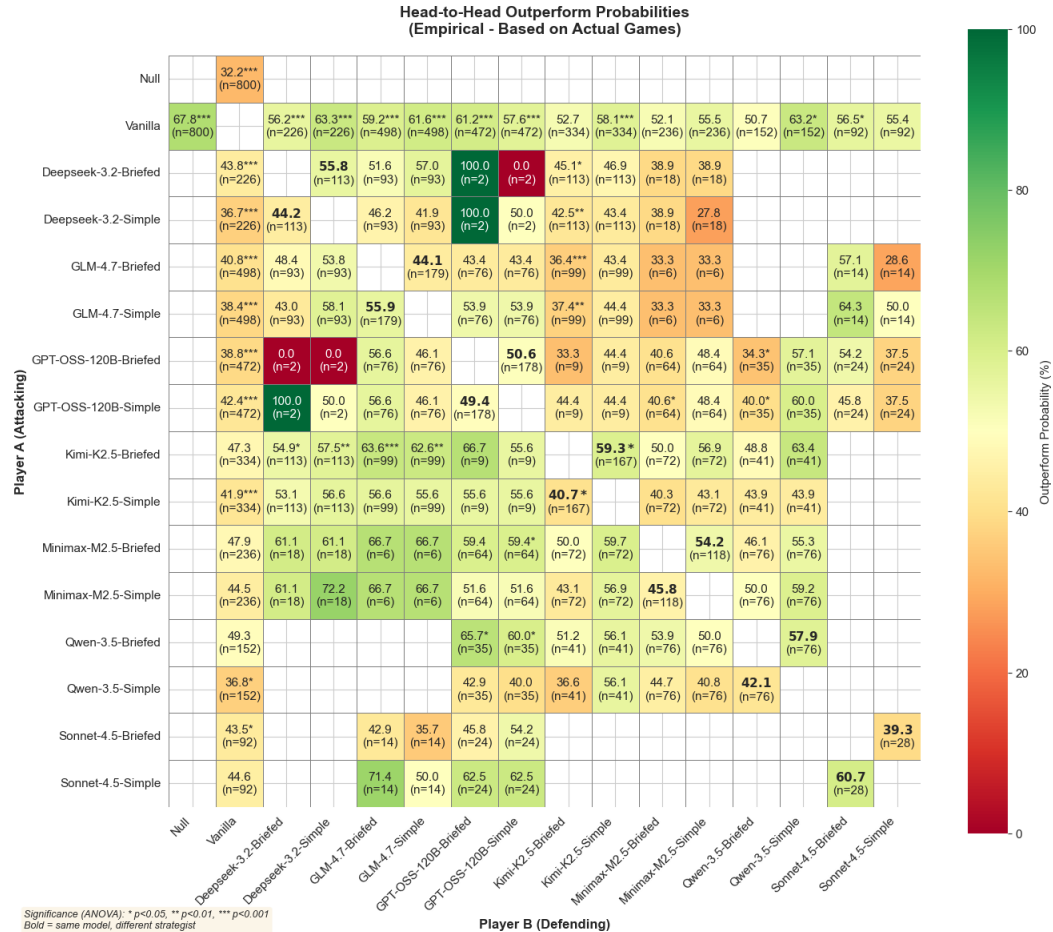

\centering
\IfFileExists{notebooks/extracted/performance/turn\_predicted/images/cell\_12\_out\_1.png}{%
\includegraphics[width=1\linewidth]{notebooks/extracted/performance/turn\_predicted/images/cell\_12\_out\_1.png}%
}{%
\includegraphics[width=1.2\linewidth]{notebooks/extracted/performance/turn\_predicted/images/cell\_13\_out\_1.png}%
}
\caption{Empirical head-to-head outperform probabilities.}
\label{fig:appendix-head-to-head}
\end{figure}

\subsubsection{Strategic Profile Diagnostics}

\begin{figure}[H]
\centering
\includegraphics[width=0.90\linewidth]{notebooks/extracted/profiles/victory\_commitment/images/cell\_07\_out\_1.png}
\caption{Strategic commitment relative to Vanilla (positive = more concentrated).}
\label{fig:appendix-strategic-commitment}
\end{figure}

\begin{figure}[H]
\centering
\includegraphics[width=0.58\linewidth]{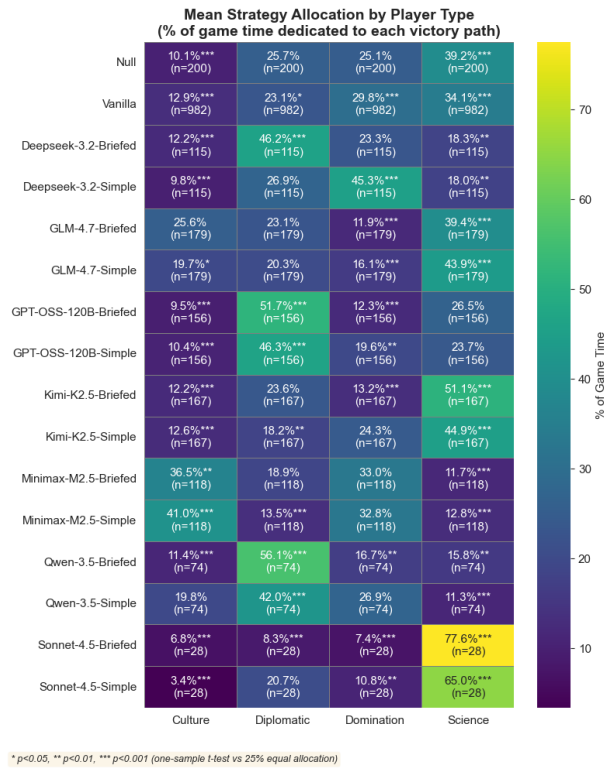}
\caption{Strategy time allocation (mean share of game time) by player type.}
\label{fig:strategy-time-allocation}
\end{figure}

\begin{figure}[!htbp]
\centering
\includegraphics[width=1\linewidth]{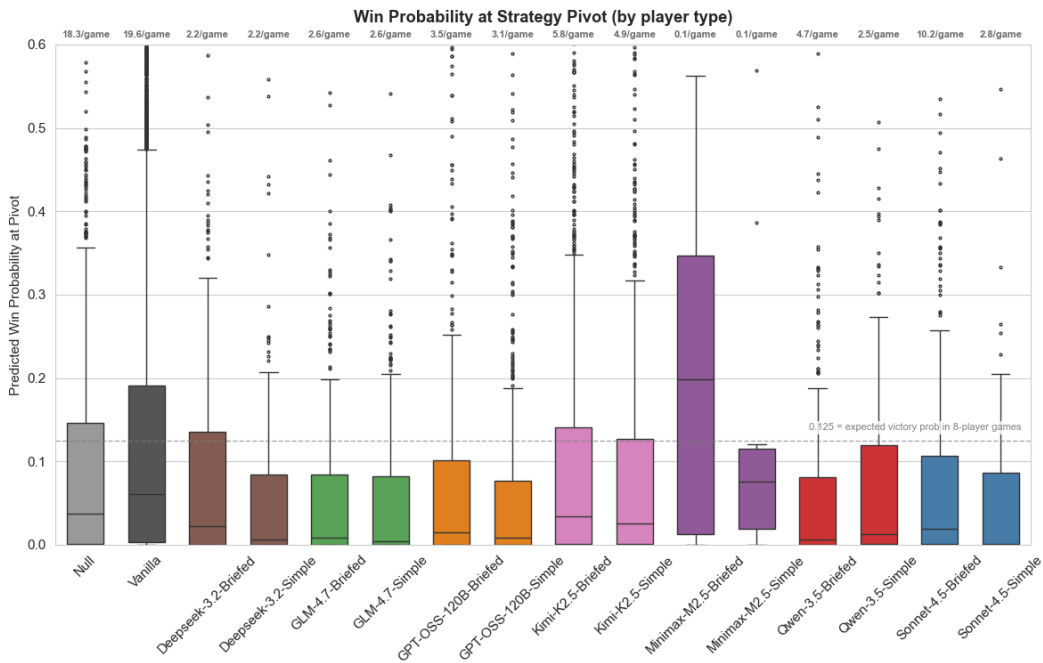}
\caption{Win probability at strategy pivot.}
\label{fig:win-probability-at-pivot}
\end{figure}

\begin{figure}[H]
\centering
\includegraphics[width=1\linewidth]{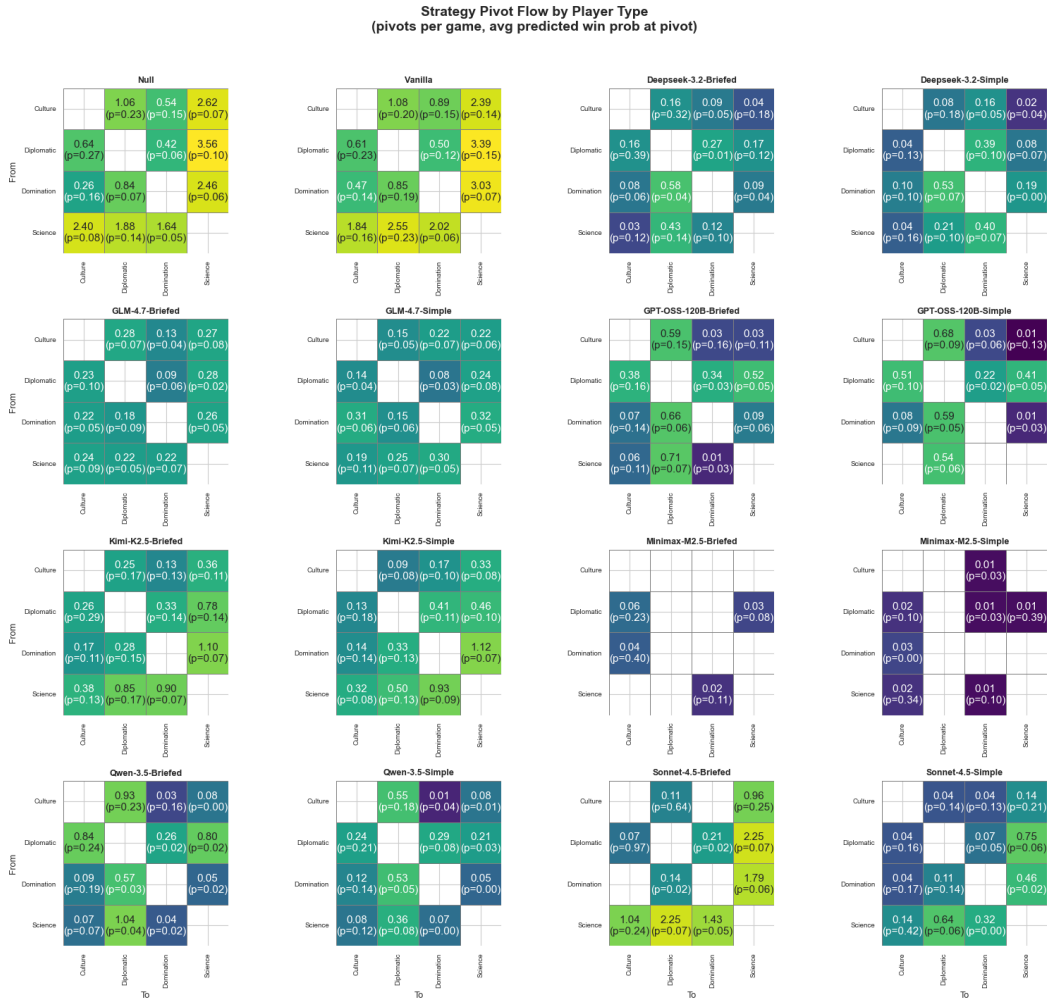}
\caption{Pivot-flow heatmaps by player type. Each facet shows from$\rightarrow$to transition rates per game, with average win probability at pivot annotated in-cell.}
\label{fig:appendix-pivot-flow-heatmaps}
\end{figure}

\subsubsection{Bradley-Terry Rating Convergence Ablation}
\label{sec:appendix-bt-convergence}

To assess the stability of Bradley-Terry ratings given the study's sample size, we conducted a convergence ablation for each player type. For a target player type X with N games sorted chronologically, we hold all games \emph{not} containing X as a constant base set (~300 games), then progressively add X's games one at a time (steps k = 1, ..., N), refitting the Bradley-Terry model at each step. This isolates X's marginal evidence contribution: all background ratings remain stable from the base set, and only X's rating moves meaningfully as its games accumulate.

\begin{figure}[H]
\centering
\includegraphics[width=0.90\linewidth]{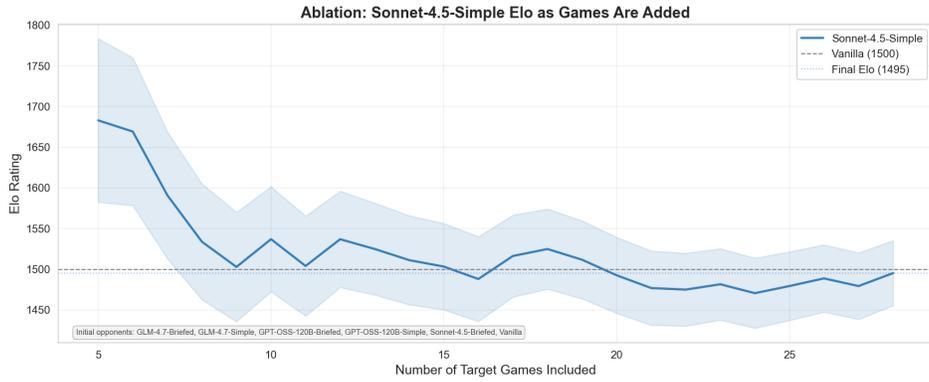}
\caption{Ablation: Sonnet-4.5-Simple}
\label{fig:appendix-ablation-sonnet}
\end{figure}

\begin{figure}[H]
\centering
\includegraphics[width=0.90\linewidth]{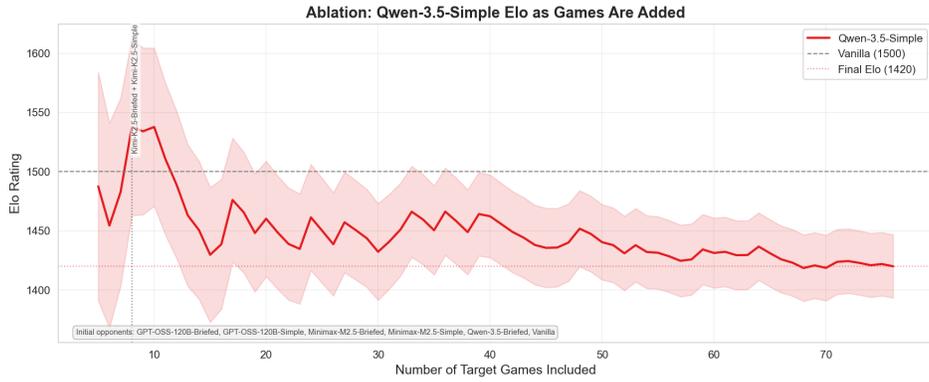}
\caption{Ablation: Qwen-3.5-Simple}
\label{fig:appendix-ablation-qwen}
\end{figure}

\begin{figure}[H]
\centering
\includegraphics[width=0.90\linewidth]{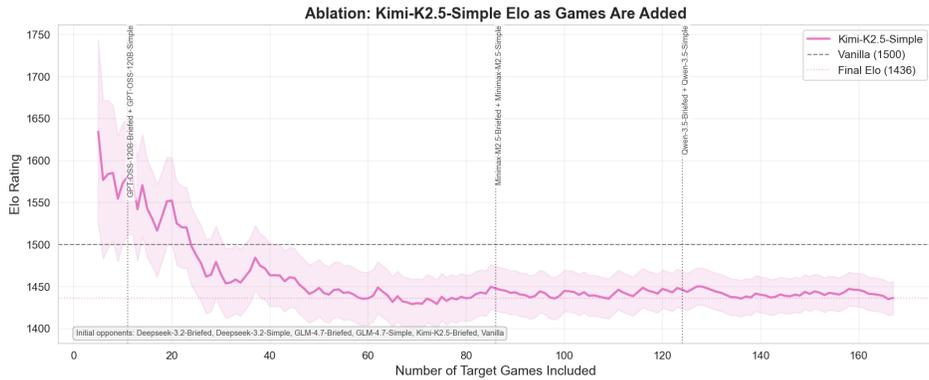}
\caption{Ablation: Kimi-K2.5-Simple}
\label{fig:appendix-ablation-kimi}
\end{figure}

The ablation reveals three contrasting convergence patterns (Appendix Figures~\ref{fig:appendix-ablation-sonnet}--\ref{fig:appendix-ablation-kimi}). Sonnet-4.5-Simple converges rapidly to near-VPAI level (~1503) within 20 games with narrow confidence intervals, indicating consistent cross-matchup performance (Appendix Figure~\ref{fig:appendix-ablation-sonnet}). Qwen-3.5-Simple illustrates the opposite risk: a few strong early performances inflate the initial estimate to ~1550 before it corrects to ~1450 around 20 games, dropping to ~1425 range at 50-60 games (Appendix Figure~\ref{fig:appendix-ablation-qwen}). Kimi-K2.5-Simple is volatile before ~40 games but eventually stabilizes around its final value (1436) at 50-60 games (Appendix Figure~\ref{fig:appendix-ablation-kimi}). These patterns suggest the corpus is sufficient to separate broad tiers, while also surfacing which player types would benefit most from additional data.

\subsubsection{Bootstrap Uncertainty for Bradley-Terry Ratings}
\label{sec:appendix-bootstrap}

To quantify how the unbalanced schedule affects the reported rankings, we ran a game-level nonparametric bootstrap over the full corpus. Each replicate resamples games with replacement and refits the Bradley-Terry model (B = 2000). We ran two variants: a stratified bootstrap, which resamples games within each experiment to preserve the schedule, and a non-stratified bootstrap, which resamples from the full game pool. The pairwise margin threshold is computed once on the original data and frozen across all replicates, and we report percentile confidence intervals around the full-sample point estimates. Because each replicate refits the model under this frozen-margin convention, the bootstrap point estimates can differ marginally from the main-text ELO values.

Figure~\ref{fig:appendix-bootstrap-forest} shows per-player-type rating intervals, and Figure~\ref{fig:appendix-bootstrap-dominance} reports the pairwise dominance matrix: the fraction of replicates in which the row player's rating exceeds the column player's. We found clear separation between broad tiers, while adjacent rankings should not be over-read. Table~\ref{tab:appendix-bootstrap-pairs} highlights representative pairs, and the two variants agree closely. Nearby orderings remain uncertain, with dominance close to 0.5: for example, Kimi-K2.5-Briefed exceeds Vanilla in only 52\% of stratified replicates (+3 ELO). Tier separations are stable, with dominance at or above roughly 0.9: for example, Qwen-3.5-Briefed exceeds Kimi-K2.5-Simple in 95\% (+60 ELO). 

\begin{table}[H]
\centering
\caption{Bootstrap support for representative pairwise ordering claims. Support is the fraction of replicates in which the first player's Bradley-Terry rating exceeds the second's.}
\label{tab:appendix-bootstrap-pairs}
\small
\begin{adjustbox}{max width=\linewidth}
\begin{tabular}{l r r r}
\toprule
Pairwise claim & ELO diff & Stratified support & Non-stratified support \\
\midrule
Kimi-K2.5-Briefed $>$ Vanilla & +3 & 0.52 & 0.55 \\
Sonnet-4.5-Simple $>$ Minimax-M2.5-Briefed & +6 & 0.56 & 0.53 \\
Minimax-M2.5-Simple $>$ Kimi-K2.5-Simple & +33 & 0.84 & 0.84 \\
Sonnet-4.5-Briefed $>$ Null & +59 & 0.84 & 0.84 \\
Sonnet-4.5-Simple $>$ GPT-OSS-120B-Simple & +82 & 0.92 & 0.90 \\
Qwen-3.5-Briefed $>$ Kimi-K2.5-Simple & +60 & 0.95 & 0.93 \\
\bottomrule
\end{tabular}
\end{adjustbox}
\end{table}

\begin{figure}[H]
\centering
\includegraphics[width=0.85\linewidth]{notebooks/bootstrap/bootstrap\_bt\_forest.png}
\caption{Stratified bootstrap (B = 2000) percentile intervals for Bradley-Terry ratings by player type.}
\label{fig:appendix-bootstrap-forest}
\end{figure}

\begin{figure}[H]
\centering
\includegraphics[width=0.9\linewidth]{notebooks/bootstrap/bootstrap\_bt\_dominance.png}
\caption{Stratified bootstrap pairwise dominance matrix. Each cell shows the fraction of replicates in which the row player's rating exceeds the column player's.}
\label{fig:appendix-bootstrap-dominance}
\end{figure}

\subsubsection{LLM Inference Costs}
\label{sec:appendix-llm-inference-costs}
Cost estimation is derived from OpenRouter as of Mar 26, 2026 and does not include input cache or batch inference discounts. Total = \$10{,}497.

\begin{table}[H]
\centering
\caption{LLM inference costs by strategist-model pairing.}
\label{tab:appendix-llm-inference-costs}
\small
\begin{adjustbox}{max width=\linewidth}
\begin{tabular}{l l r l l l l}
\toprule
Strategist & Model Name & Games & Avg. Input & Avg. Output & Cost Per Unit & Total Cost \\
\midrule
Deepseek-3.2-Briefed & Deepseek-3.2 & 113 & 7,109,358 & 817,451 & \$0.26 in / \$0.38 out per 1M & \$243.97 \\
Deepseek-3.2-Briefed & GPT-OSS-120B & 113 & 15,162,780 & 1,987,563 & \$0.04 in / \$0.19 out per 1M & \$109.50 \\
Deepseek-3.2-Simple & Deepseek-3.2 & 113 & 14,349,173 & 725,711 & \$0.26 in / \$0.38 out per 1M & \$452.74 \\
GLM-4.7-Briefed & GLM-4.7 & 178 & 7,599,886 & 911,523 & \$0.39 in / \$1.75 out per 1M & \$811.52 \\
GLM-4.7-Briefed & GPT-OSS-120B & 178 & 15,737,605 & 2,354,194 & \$0.04 in / \$0.19 out per 1M & \$188.87 \\
GLM-4.7-Simple & GLM-4.7 & 179 & 13,725,978 & 825,193 & \$0.39 in / \$1.75 out per 1M & \$1,216.70 \\
GPT-OSS-120B-Briefed & GPT-OSS-120B & 145 & 24,887,643 & 3,138,016 & \$0.04 in / \$0.19 out per 1M & \$227.19 \\
GPT-OSS-120B-Simple & GPT-OSS-120B & 145 & 13,774,302 & 603,550 & \$0.04 in / \$0.19 out per 1M & \$94.52 \\
Kimi-K2.5-Briefed & GPT-OSS-120B & 167 & 16,934,443 & 2,120,138 & \$0.04 in / \$0.19 out per 1M & \$177.57 \\
Kimi-K2.5-Briefed & Kimi-K2.5 & 167 & 7,481,619 & 1,113,706 & \$0.45 in / \$2.20 out per 1M & \$971.42 \\
Kimi-K2.5-Simple & Kimi-K2.5 & 167 & 13,841,556 & 887,317 & \$0.45 in / \$2.20 out per 1M & \$1,366.19 \\
Minimax-M2.5-Briefed & GPT-OSS-120B & 118 & 17,410,844 & 2,498,385 & \$0.04 in / \$0.19 out per 1M & \$136.14 \\
Minimax-M2.5-Briefed & Minimax-M2.5 & 118 & 12,536,785 & 533,287 & \$0.20 in / \$1.17 out per 1M & \$369.49 \\
Minimax-M2.5-Simple & Minimax-M2.5 & 118 & 18,202,331 & 433,719 & \$0.20 in / \$1.17 out per 1M & \$489.45 \\
Qwen-3.5-Briefed & GPT-OSS-120B & 76 & 16,604,517 & 2,269,583 & \$0.04 in / \$0.19 out per 1M & \$81.99 \\
Qwen-3.5-Briefed & Qwen-3.5 & 76 & 7,436,317 & 1,430,358 & \$0.39 in / \$2.34 out per 1M & \$474.79 \\
Qwen-3.5-Simple & Qwen-3.5 & 76 & 13,370,294 & 1,283,402 & \$0.39 in / \$2.34 out per 1M & \$624.54 \\
Sonnet-4.5-Briefed & GPT-OSS-120B & 28 & 12,298,155 & 1,642,895 & \$0.04 in / \$0.19 out per 1M & \$22.17 \\
Sonnet-4.5-Briefed & Sonnet-4.5 & 28 & 5,323,294 & 1,306,414 & \$3.00 in / \$15.00 out per 1M & \$995.85 \\
Sonnet-4.5-Simple & Sonnet-4.5 & 28 & 12,415,136 & 951,373 & \$3.00 in / \$15.00 out per 1M & \$1,442.45 \\
\bottomrule
\end{tabular}
\end{adjustbox}
\end{table}

\subsubsection{Briefer Instruction Intensity}

To quantify how actively each briefed strategist manages the fixed GPT-OSS-120B briefing pipeline, we aggregate \texttt{focus\_briefer\_count} per game across all strategist turns. Higher counts indicate that the strategist more frequently used the \texttt{focus-briefer} tool to redirect the Military, Economy, or Diplomacy briefers.

\begin{figure}[H]
\centering
\includegraphics[width=0.90\linewidth]{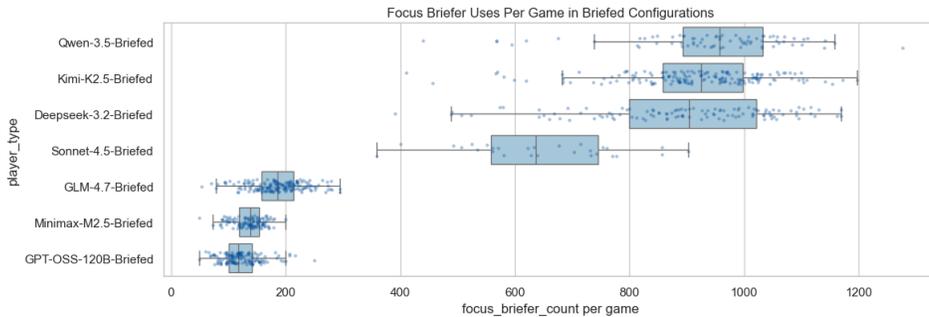}
\caption{Distribution of \texttt{focus\_briefer\_count} per game for briefed configurations.}
\label{fig:appendix-briefer-instruction-intensity}
\end{figure}

\subsection{Logged Signal Definitions}
\label{sec:appendix-logged-signals}

CivBench logs two classes of signals: \textbf{game-state signals} recorded each turn from the observable game state, and \textbf{agentic decision signals} recorded when the LLM makes a decision.

\subsubsection{Game-state Signals}

\textbf{\texttt{PlayerSummaries}}: Per-player per-turn snapshot of competitive standing and resource state.

\begin{table}[H]
\centering
\small
\begin{adjustbox}{max width=\linewidth}
\begin{tabular}{l l}
\toprule
Field & Description \\
\midrule
Score & In-game civilization score \\
Era & Current historical era (e.g., Ancient, Medieval, Industrial) \\
MajorAlly & Allied major civilization, if any \\
Votes & United Nations vote count \\
Cities & Number of cities controlled \\
Population & Total population across all cities \\
Territory & Number of map plots owned \\
Gold & Current gold treasury \\
GoldPerTurn & Net gold income per turn \\
HappinessPercentage & Excess happiness percentage (can be negative) \\
GoldenAge & Golden Age status (e.g., "5 turns remaining", "Need More Happiness") \\
SciencePerTurn & Science output per turn \\
CulturePerTurn & Culture output per turn \\
FaithPerTurn & Faith output per turn \\
TourismPerTurn & Tourism output per turn \\
PolicyBranches & JSON: policy branch name $\rightarrow$ adoption status \\
Technologies & Count of technologies researched \\
CurrentResearch & Technology currently being researched \\
NextPolicyTurns & Turns until next policy can be adopted \\
MilitaryUnits & Current military units in supply \\
MilitarySupply & Maximum military supply capacity \\
MilitaryStrength & Total military strength (sum of unit attack power) \\
ResourcesAvailable & Strategic and luxury resources available \\
FoundedReligion & Religion founded by this player, if any \\
MajorityReligion & Religion with majority presence in player's cities \\
Relationships & JSON: diplomatic stance toward each other player \\
OutgoingTradeRoutes & Active outgoing trade routes \\
IncomingTradeRoutes & Active incoming trade routes \\
Spies & Spy deployment information \\
DiplomaticDeals & Active diplomatic agreements \\
Quests & Active city-state quests \\
BestSettlementLocation & VPAI's recommended next settlement in revealed map \\
\bottomrule
\end{tabular}
\end{adjustbox}
\end{table}

\textbf{\texttt{VictoryProgress}}: Per-player per-turn progress toward each victory type.

\begin{table}[H]
\centering
\small
\begin{adjustbox}{max width=\linewidth}
\begin{tabular}{l l}
\toprule
Field & Description \\
\midrule
DominationVictory & Progress status toward Domination victory (JSON) \\
ScienceVictory & Progress status toward Science victory (JSON) \\
CulturalVictory & Progress status toward Cultural victory (JSON) \\
DiplomaticVictory & Progress status toward Diplomatic victory (JSON) \\
\bottomrule
\end{tabular}
\end{adjustbox}
\end{table}

\textbf{\texttt{CityInformations}}: Per-city per-turn data at two visibility levels.

\emph{Basic fields (visible to all players):}

\begin{table}[H]
\centering
\small
\begin{adjustbox}{max width=\linewidth}
\begin{tabular}{l l}
\toprule
Field & Description \\
\midrule
Owner & Owning civilization \\
Name & City name \\
X, Y & Map coordinates \\
Population & City population \\
MajorityReligion & Dominant religion in the city \\
DefenseStrength & City defense strength \\
HitPoints / MaxHitPoints & Current and maximum city HP \\
IsCapital / IsPuppet / IsOccupied / IsCoastal & City status flags \\
\bottomrule
\end{tabular}
\end{adjustbox}
\end{table}

\emph{Full fields (own and fully visible cities only):}

\begin{table}[H]
\centering
\small
\begin{tabularx}{\textwidth}{>{\raggedright\arraybackslash}p{0.4\textwidth} X}
\toprule
Field & Description \\
\midrule
FoodStored / FoodPerTurn & Food stored and net food income \\
ProductionStored / ProductionPerTurn & Production stored and output \\
GoldPerTurn / SciencePerTurn / CulturePerTurn / FaithPerTurn / TourismPerTurn & Per-turn yields \\
HappinessDelta & Happiness contribution \\
RazingTurns / ResistanceTurns & Turns until razing completes / resistance ends \\
BuildingCount & Number of buildings \\
Wonders & JSON array of wonders present \\
ImportantBuildings & JSON array of notable buildings \\
GreatWorkCount & Number of Great Works housed \\
CurrentProduction / ProductionTurnsLeft & What is being built and turns to completion \\
\bottomrule
\end{tabularx}
\end{table}

\textbf{\texttt{TacticalZones}}: Per-player per-turn military positioning data, as analyzed by VPAI's tactical module within the fog of war.

\begin{table}[H]
\centering
\small
\begin{tabularx}{\textwidth}{>{\raggedright\arraybackslash}p{0.4\textwidth} X}
\toprule
Field & Description \\
\midrule
PlayerID & Player this zone belongs to \\
ZoneID & Zone identifier \\
Territory & Ownership type of the zone \\
Dominance & Military control (friendly / contested / enemy) \\
Domain & Combat domain (land / sea / air) \\
Posture & Military posture (defensive / offensive / etc.) \\
AreaID & Geographic area identifier \\
City & Associated city, if any \\
CenterX / CenterY & Zone center coordinates \\
Plots & Number of plots in the zone \\
Value & Strategic value of the zone \\
FriendlyStrength / EnemyStrength / NeutralStrength & Military strength by side \\
Neighbors & JSON array of adjacent zone IDs \\
Units & JSON: civilization $\rightarrow$ unit type $\rightarrow$ count \\
\bottomrule
\end{tabularx}
\end{table}

\textbf{\texttt{RelationshipChanges}}: Diplomatic relationship updates, logged when the LLM updates its stance toward another player.

\begin{table}[H]
\centering
\small
\begin{adjustbox}{max width=\linewidth}
\begin{tabular}{l l}
\toprule
Field & Description \\
\midrule
PlayerID & The agent recording the relationship \\
TargetID & The target player \\
PublicValue & Declared diplomatic stance (0--100) \\
PrivateValue & Internal (hidden) stance (0--100) \\
Rationale & LLM's written rationale for the stance \\
\bottomrule
\end{tabular}
\end{adjustbox}
\end{table}

\textbf{\texttt{GameEvents}}: Detailed records of in-game events each turn (e.g., declarations of war, city captures, wonder completions, great person births). Columns include Type and Turn, plus all event-specific detail fields.

\subsubsection{Agentic Decision Signals}

These tables record what the LLM chose, not what happened in the game state. Each entry is logged at the time of the decision.

\textbf{\texttt{PolicyChanges}}: LLM's policy adoption decisions.

\begin{table}[H]
\centering
\small
\begin{adjustbox}{max width=\linewidth}
\begin{tabular}{l l}
\toprule
Field & Description \\
\midrule
Policy & Policy or policy branch name \\
IsBranch & Whether this is a branch adoption (1) or individual policy (0) \\
Rationale & LLM's written rationale \\
\bottomrule
\end{tabular}
\end{adjustbox}
\end{table}

\textbf{\texttt{ResearchChanges}}: LLM's technology research decisions.

\begin{table}[H]
\centering
\small
\begin{adjustbox}{max width=\linewidth}
\begin{tabular}{l l}
\toprule
Field & Description \\
\midrule
Technology & Technology name chosen for research \\
Rationale & LLM's written rationale \\
\bottomrule
\end{tabular}
\end{adjustbox}
\end{table}

\textbf{\texttt{PersonaChanges}}: LLM's diplomatic personality parameter settings (integer 0--10). These values bias Vox Populi AI's diplomatic behavior.

\emph{Competitiveness \& Ambition:}

\begin{table}[H]
\centering
\small
\begin{adjustbox}{max width=\linewidth}
\begin{tabular}{l l}
\toprule
Field & Description \\
\midrule
VictoryCompetitiveness & Drive to pursue victory conditions \\
WonderCompetitiveness & Drive to build wonders \\
MinorCivCompetitiveness & Drive to compete for city-state influence \\
Boldness & Willingness to take aggressive actions \\
\bottomrule
\end{tabular}
\end{adjustbox}
\end{table}

\emph{War \& Peace:}

\begin{table}[H]
\centering
\small
\begin{adjustbox}{max width=\linewidth}
\begin{tabular}{l l}
\toprule
Field & Description \\
\midrule
WarBias & General bias toward initiating war \\
HostileBias & Tendency toward hostile diplomatic stance \\
WarmongerHate & Sensitivity to warmonger penalties \\
NeutralBias & Tendency toward neutral stance \\
FriendlyBias & Tendency toward friendly stance \\
GuardedBias & Tendency toward guarded stance \\
AfraidBias & Tendency to act from fear \\
\bottomrule
\end{tabular}
\end{adjustbox}
\end{table}

\emph{Diplomacy \& Cooperation:}

\begin{table}[htbp]
\centering
\small
\begin{adjustbox}{max width=\linewidth}
\begin{tabular}{l l}
\toprule
Field & Description \\
\midrule
DiplomaticBalance & Balance between cooperation and competition \\
Friendliness & General friendliness level \\
WorkWithWillingness & Willingness to form alliances \\
WorkAgainstWillingness & Willingness to coordinate against others \\
Loyalty & Loyalty to existing agreements \\
\bottomrule
\end{tabular}
\end{adjustbox}
\end{table}

\emph{Minor Civ Relations:}

\begin{table}[H]
\centering
\small
\begin{adjustbox}{max width=\linewidth}
\begin{tabular}{l l}
\toprule
Field & Description \\
\midrule
MinorCivFriendlyBias / MinorCivNeutralBias / MinorCivHostileBias / MinorCivWarBias & Stances toward city-states \\
\bottomrule
\end{tabular}
\end{adjustbox}
\end{table}

\emph{Personality Traits:}

\begin{table}[H]
\centering
\small
\begin{adjustbox}{max width=\linewidth}
\begin{tabular}{l l}
\toprule
Field & Description \\
\midrule
DenounceWillingness & Willingness to publicly denounce others \\
Forgiveness & Willingness to forgive past grievances \\
Meanness & General meanness in interactions \\
Neediness & Tendency to seek reassurance or deals \\
Chattiness & Frequency of initiating diplomacy \\
DeceptiveBias & Tendency toward deceptive behavior \\
\bottomrule
\end{tabular}
\end{adjustbox}
\end{table}

\textbf{\texttt{FlavorChanges}}: LLM's strategic preference (flavor) settings, which directly bias Vox Populi AI's tactical decision-making algorithms. Also records the overall GrandStrategy and a written rationale.

\emph{Military flavors (18):} Offense, Defense, Mobilization, CityDefense, MilitaryTraining, Recon, Ranged, Mobile, Nuke, UseNuke, Naval, NavalRecon, NavalGrowth, NavalTileImprovement, Air, AirCarrier, Antiair, Airlift

\emph{Economy flavors (9):} Expansion, Growth, TileImprovement, Infrastructure, Production, WaterConnection, Gold, Science, Culture

\emph{Development flavors (7):} Happiness, GreatPeople, Wonder, Religion, Diplomacy, Spaceship, Espionage

Additional fields: GrandStrategy (overall strategic orientation), Rationale (LLM's written rationale).

\subsection{Strategist Agent Configurations}
\label{sec:appendix-agent-configurations}

This section details the two strategist configurations used in our experiments. Both share the same tool interface (the strategic controls described in the main text) and access the same underlying game-state data. They differ in how that data is presented to the strategist: directly (Simple) or via a summarization pipeline (Briefed).

\subsubsection{Shared Components}

Both configurations share the following system prompt structure:

\begin{table}[H]
\centering
\small
\begin{tabularx}{\textwidth}{l X}
\toprule
Component & Content \\
\midrule
Role & Expert player of Civilization V with Vox Populi mod \\
Expectation & Delegates all tactical decisions to in-game AI; focuses on macro-level strategy; interacts only through tool calls (no user interaction); may invoke multiple tools per turn \\
Goals & Call tools to set strategic parameters with written rationale for each decision \\
Decision constraint & Must call either \texttt{set-strategy}/\texttt{set-flavors} or \texttt{keep-status-quo} each turn \\
\bottomrule
\end{tabularx}
\end{table}

\textbf{Tools available to both configurations:}

\begin{table}[H]
\centering
\small
\begin{tabularx}{\textwidth}{l >{\raggedright\arraybackslash}p{0.3\textwidth} X}
\toprule
Tool & Parameters & Description \\
\midrule
\texttt{set-flavors} & Grand strategy, 34 flavor weights & Set overall strategic direction and tactical flavor biases \\
\texttt{set-persona} & 26 diplomatic weights (0--10) & Set diplomatic personality parameters \\
\texttt{set-relationship} & Target player, bias (--100 to 100) & Set relationship modifier toward a specific major civilization \\
\texttt{set-research} & Technology name & Choose next technology to research \\
\texttt{set-policy} & Policy name & Choose next social policy to adopt \\
\texttt{keep-status-quo} & (none) & Maintain current strategic settings unchanged \\
\bottomrule
\end{tabularx}
\end{table}

The agent loop terminates after \texttt{set-strategy}/\texttt{set-flavors} or \texttt{keep-status-quo} is called, or after 5 agentic steps, whichever comes first.

\subsubsection{Simple Configuration}

The Simple strategist receives the full game-state report each turn as a structured Markdown document:

\begin{table}[H]
\centering
\small
\begin{tabularx}{\textwidth}{l X}
\toprule
Input section & Content \\
\midrule
Situation & Turn number, era, civilization traits \\
Civilization & Owned resources, yields, happiness \\
Options & Available grand strategies, economic strategies, military strategies, flavor weights \\
Strategies & Current strategic settings and rationale from the previous turn \\
Victory Progress & Progress toward each victory condition \\
Players & Full diplomatic and competitive summary for all known players (opinion scores, relationships, alliances, deals, spy activity, military strength, economic output, era, etc.) \\
Cities & Per-city details for all discovered cities: population, production queue, buildings, defense, religion, growth \\
Military & Tactical zones with per-zone dominance, posture, value, and unit counts; military supply utilization \\
Events & All game events from the previous turn (wars declared, cities captured, wonders built, technologies researched, etc.) \\
\bottomrule
\end{tabularx}
\end{table}

The Simple configuration has no additional tools beyond the shared set. Its system prompt describes each input section to orient the model.

\subsubsection{Briefed Configuration}

In the Briefed configuration, the raw Cities, Military, and Events data are routed through a pipeline of subagents running on a fixed weaker model (GPT-OSS-120B), which summarize them into \textbf{structured briefings} delivered to the strategist. The configuration also adds one additional tool for steering the briefer subagent.

\textbf{Differences from Simple:}

\begin{table}[H]
\centering
\small
\begin{tabularx}{\textwidth}{l >{\raggedright\arraybackslash}p{0.23\textwidth} X}
\toprule
Aspect & Simple & Briefed \\
\midrule
Cities report & Full per-city detail & Routed to briefers; strategist receives summaries \\
Military report & Full unit/zone detail & Routed to briefers; strategist receives summaries \\
Events report & Full event log & Routed to briefers; strategist receives summaries \\
Players report & Full player data & Filtered to strategic-level fields (removes raw opinion scores and spy details) \\
Briefings section & Not present & Structured briefings from subagent pipeline \\
Additional tool & -- & \texttt{focus-briefer} \\
\bottomrule
\end{tabularx}
\end{table}

\textbf{Additional tool:}

\begin{table}[H]
\centering
\small
\begin{tabularx}{\textwidth}{l >{\raggedright\arraybackslash}p{0.4\textwidth} X}
\toprule
Tool & Parameters & Description \\
\midrule
\texttt{focus-briefer} & Mode (Military / Economy / Diplomacy), natural-language instruction & Instruct a specific briefer to focus on particular topics in next turn's report (stored in working memory) \\
\bottomrule
\end{tabularx}
\end{table}

The system prompt additionally informs the strategist of its three specialized briefers (Military, Economy, Diplomacy) and describes their limitations: briefers only see limited current game state, cannot control or predict tactical AI decisions, and are best at summarizing and synthesizing factual information rather than analyzing, projecting, or predicting.

\subsubsection{Briefing Pipeline}

The briefing pipeline runs before each strategist's turn to compress raw game state into structured briefings. It uses \textbf{conditional routing} based on context length:

\begin{itemize}
    \item \textbf{Short context} (events $\leq$ 5,000 characters OR turn $\leq$ 1): A \textbf{single briefer}---one GPT-OSS-120B call with combined Military + Economy + Diplomacy instructions---produces a single comprehensive briefing.
    \item \textbf{Long context} (otherwise): \textbf{Three specialized briefers} run as parallel GPT-OSS-120B calls, one per domain, producing three domain-specific briefings (Military, Economy, Diplomacy).
\end{itemize}

Each specialized briefer receives:
\begin{itemize}
    \item \textbf{Military briefer}:
    \begin{itemize}
        \item \emph{Players data}: Filtered (military-relevant: strength, supply, era, gold, technologies, policy branches)
        \item \emph{Cities data}: Filtered (defense-relevant: HP, defense strength, production queue)
        \item \emph{Events}: Military events only
        \item \emph{Additional data}: Full tactical zones and unit roster
    \end{itemize}

    \item \textbf{Economy briefer}:
    \begin{itemize}
        \item \emph{Players data}: Full (minus diplomatic opinion and spy fields)
        \item \emph{Cities data}: Full (minus majority religion)
        \item \emph{Events}: Economy events only
        \item \emph{Additional data}: Victory progress
    \end{itemize}

    \item \textbf{Diplomacy briefer}:
    \begin{itemize}
        \item \emph{Players data}: Filtered (diplomatic: opinions, relationships, allies, quests, deals, spies, religion)
        \item \emph{Cities data}: Filtered (religion, influence-relevant)
        \item \emph{Events}: Diplomacy events only
        \item \emph{Additional data}: World Congress data
    \end{itemize}
\end{itemize}

All briefers share these constraints (enforced via system prompt):
\begin{itemize}
\item Objective and analytical; no speculation, estimation, or strategic recommendations
\item Macro-level summaries only; no raw tactical detail (e.g., coordinates, unit IDs)
\item Never suggest actions, as the leader makes independent decisions
\item Receive past briefing from approximately 5 turns ago for temporal comparison
\end{itemize}

\subsection{Estimation Model Features and Specifications}
\label{sec:appendix-estimation-specs}

\subsubsection{Feature Definitions}

The estimation models use 23 game-state features organized into 9 strategic groups, plus \texttt{turn\_progress} (elapsed turns / max turns) as a 24th input for all models except Baseline (excluded to avoid temporal confounding in coefficient estimation). For yield-type dimensions, three encodings exist: \texttt{\_adj} (adjusted) values penalize raw yields for city count, reflecting Vox Populi's city-count penalty on per-city output; \texttt{\_share} values express \texttt{\_adj} as a fraction of the cross-player total; \texttt{\_raw\_share} values express the raw (unadjusted) yield as a fraction. Baseline and XGBoost use \texttt{\_share} features; MLP and GroupedMLP use \texttt{\_share} for most yields but \texttt{\_raw\_share} for production and faith; InteractionMLP and AttentionMLP use \texttt{\_adj} values directly, letting their set-based architectures learn cross-player comparisons internally.

\begin{table}[H]
\centering
\small
\begin{tabularx}{\textwidth}{l >{\raggedright\arraybackslash}p{0.4\textwidth} X}
\toprule
Group & Feature (share / absolute) & Description \\
\midrule
\textbf{Science} & \texttt{technologies\_gap} & Technologies the leader has researched that this player has not \\
 & \texttt{science\_share} / \texttt{science\_adj} & Science output per turn \\
\textbf{Culture} & \texttt{policies\_gap} & Civics/policies the leader has adopted that this player has not \\
 & \texttt{culture\_share} / \texttt{culture\_adj} & Culture output per turn \\
 & \texttt{tourism\_share} / \texttt{tourism\_adj} & Tourism generated; tracks Cultural victory progress \\
\textbf{Economy} & \texttt{gold\_share} / \texttt{gold\_adj} & Gold income per turn \\
 & \texttt{production\_raw\_share} / \texttt{production\_adj} & Production (hammers) output \\
\textbf{Growth} & \texttt{cities\_share} / \texttt{cities} & Cities controlled \\
 & \texttt{food\_share} / \texttt{food\_adj} & Food output; city growth potential \\
 & \texttt{population\_share} / \texttt{population} & Total population across player's cities \\
\textbf{Religion} & \texttt{faith\_raw\_share} / \texttt{faith\_adj} & Faith output per turn \\
 & \texttt{religion\_percentage} & Fraction of world cities following player's religion \\
\textbf{Influence} & \texttt{votes\_share} / \texttt{votes} & World Congress votes; Diplomatic victory track \\
 & \texttt{minor\_allies\_share} / \texttt{minor\_allies} & City-state suzerain relationships \\
 & \texttt{defensive\_pacts} & Mutual defense agreements in effect \\
 & \texttt{friendships} & Declared Friendship agreements with major players \\
\textbf{War} & \texttt{military\_share} / \texttt{military\_adj} & Military unit strength \\
 & \texttt{military\_utilization} & Fraction of military supply in use \\
 & \texttt{active\_wars} & Wars currently engaged in \\
 & \texttt{truces} & Active truces (recently ended wars) \\
\textbf{Welfare} & \texttt{happiness\_percentage} & Fraction of population that is content (sufficient amenities) \\
 & \texttt{highest\_war\_weariness} & Maximum war weariness level \\
\textbf{Score} & \texttt{score\_ratio} & Player's score as share of total score across all players \\
\emph{(Temporal)} & \texttt{turn\_progress}* & Elapsed turns / max turns \\
\bottomrule
\end{tabularx}
\end{table}

* Baseline excludes \texttt{turn\_progress} (23 features); all other models include it (24 inputs). Baseline and XGBoost feature pipelines report \texttt{production\_share} and \texttt{faith\_share} (city-adjusted shares), which differ from the \texttt{production\_raw\_share} and \texttt{faith\_raw\_share} (raw-total shares) used by MLP and GroupedMLP.

\subsubsection{Model Specifications}

All neural models use GELU activations, LayerNorm, residual connections (when depth $\geq$ 2 layers), and mixed-precision training (AMP) on CUDA. Hyperparameters were tuned via Optuna \citep{Optuna} (1,000 trials each). Each neural model's per-sample loss is weighted by \texttt{turn\_progress}\^{\texttt{loss\_tp\_alpha}}, up-weighting later-game observations where outcomes are more certain; \texttt{loss\_tp\_alpha} = 0 recovers uniform weighting.

\begin{table}[H]
\centering
\small
\begin{tabularx}{\textwidth}{l >{\raggedright\arraybackslash}p{0.3\textwidth} >{\raggedright\arraybackslash}p{0.2\textwidth} X}
\toprule
Model & Architecture & Hyperparameters & Training \\
\midrule
\textbf{Baseline} & Logistic regression (statsmodels Logit); standardized features; cluster-robust SEs by game\_id & 23 features (no \texttt{turn\_progress}) & Isotonic calibration via 5-fold OOF predictions \\
\textbf{XGBoost} & XGBClassifier; \texttt{scale\_pos\_weight} for class imbalance & 100 trees, max\_depth=3, lr=0.189, subsample=0.910, colsample=0.822, $\lambda$=16.89 & Two-stage: early stopping (10 rounds, 10\% holdout), then CalibratedClassifierCV (isotonic, cv=5) \\
\textbf{MLP} & \texttt{\_UtilityNet}: 1 hidden layer (60 units), GELU & \texttt{loss\_tp\_alpha}=0.414, dropout=0.448 & 27 epochs, batch=32768, Adam (lr=2.75e-3, wd=2.80e-4) \\
\textbf{GroupedMLP} & \texttt{\_UtilityNet}: 1 hidden layer (222 units), GELU; softmax over (game\_id, turn) groups & Group-wise CE; \texttt{loss\_tp\_alpha}=0.398; dropout=0.211 & 21 epochs, batch=4096 groups, Adam (lr=7.78e-3, wd=3.11e-3) \\
\textbf{InteractionMLP} & DeepSets: shared encoder (80 units) $\rightarrow$ mean+max pool $\rightarrow$ concat $\rightarrow$ shared decoder (183 units) $\rightarrow$ group softmax & \texttt{loss\_tp\_alpha}=0.104, dropout=0.322 & 29 epochs, Adam (lr=2.60e-4, wd=7.39e-4) \\
\textbf{AttentionMLP} & Shared encoder (30 units) $\rightarrow$ 1-layer pre-norm MHA (3 heads, attn\_dropout=0.259) + residual $\rightarrow$ shared decoder (25 units) $\rightarrow$ group softmax & \texttt{loss\_tp\_alpha}=0.048, dropout=0.261 & 13 epochs, Adam (lr=3.03e-3, wd=4.51e-3) \\
\bottomrule
\end{tabularx}
\end{table}

\end{document}